\documentclass[11pt,a4paper]{article}
\usepackage{amsfonts,amssymb,amsmath,mathrsfs,graphicx,xcolor,bm,mathtools,physics,orcidlink,url}
\usepackage{authblk,arydshln,subcaption,fullpage}
\usepackage{pdflscape}
\definecolor{ultramarine}{rgb}{0.07, 0.04, 0.56}
\definecolor{cadmiumgreen}{rgb}{0.0, 0.42, 0.24}
\definecolor{indigo(dye)}{rgb}{0.0, 0.25, 0.42}
\usepackage{hyperref}
\hypersetup{
colorlinks=true,
citecolor=ultramarine,
linkcolor=cadmiumgreen,
urlcolor=indigo(dye),
}

\usepackage[
  backend=biber,
  style=numeric,
  sorting=none,
  sortcites=true,
  natbib=true,
  doi=false,
  url=false,
  isbn=false,
  eprint=true,
  giveninits=true
]{biblatex}
\DeclareFieldFormat[article]{title}{\mkbibquote{#1}}
\DeclareFieldFormat{pages}{#1}
\DeclareFieldFormat{eprint:arxiv}{%
  \mkbibbrackets{\href{https://arxiv.org/abs/#1}{arXiv\addcolon#1}}}
\DeclareBibliographyDriver{article}{%
  \usebibmacro{bibindex}%
  \usebibmacro{begentry}%
  \printnames{author}%
  \setunit{\labelnamepunct}\newblock
  \printfield[title]{title}%
  \newunit\newblock
  \iffieldundef{doi}
    {%
      \printfield{journaltitle}%
      \setunit{\addspace}%
      \printfield{volume}%
      \setunit{\addcomma\space}%
      \iffieldundef{eid}
        {\iffieldundef{pages}{}{\printfield{pages}}}
        {\printfield{eid}}%
      \setunit{\addspace}%
      \printtext[parens]{\printfield{year}}%
    }
    {%
      \href{https://doi.org/\thefield{doi}}{%
        \printfield{journaltitle}%
        \setunit{\addspace}%
        \printfield{volume}%
        \setunit{\addcomma\space}%
        \iffieldundef{eid}
          {\iffieldundef{pages}{}{\printfield{pages}}}
          {\printfield{eid}}%
        \setunit{\addspace}%
        \printtext[parens]{\printfield{year}}%
      }%
    }%
  \newunit\newblock
  \printfield[eprint:arxiv]{eprint}%
  \usebibmacro{finentry}}

\begin{document}

\title{Autoencoder-Based Parameter Estimation for Superposed Multi-Component Damped Sinusoidal Signals}

\author[1,2]{Momoka Iida\orcidlink{0009-0008-7811-9446}\footnote{Corresponding author}\footnote{momoka.iida@g.kogakuin.jp}}
\author[3]{Hayato Motohashi\orcidlink{0000-0002-4330-7024}\footnote{motohashi@tmu.ac.jp}}
\author[2,4,5]{Hirotaka Takahashi\orcidlink{0000-0003-0596-4397}\footnote{hirotaka@tcu.ac.jp}}

\affil[1]{Graduate School of Integrative Science and Engineering, Tokyo City University, 1-28-1 Tamazutsumi, Setagaya-ku, Tokyo 158-8557, Japan}
\affil[2]{Research Center for Space Science, Advanced Research Laboratories, Tokyo City University, 1-28-1 Tamazutsumi, Setagaya-ku, Tokyo 158-8557, Japan}
\affil[3]{Department of Physics, Graduate School of Science, Tokyo Metropolitan University, 1-1 Minami-Osawa, Hachioji, Tokyo 192-0397, Japan}
\affil[4]{Graduate School of Information and Data Science and Department of Design and Data Science, Tokyo City University, 3-3-1 Ushikubo-Nishi, Tsuzuki-ku, Yokohama, Kanagawa 224-8551, Japan}
\affil[5]{Earthquake Research Institute, The University of Tokyo, 1-1-1 Yayoi, Bunkyo-ku, Tokyo 113-0032, Japan}

\date{}

\maketitle

\begin{abstract}
Damped sinusoidal oscillations are widely observed in many physical systems, and their analysis provides access to underlying physical properties. 
However, parameter estimation becomes difficult when the signal decays rapidly, multiple components are superposed, and observational noise is present. 
In this study, we develop an autoencoder-based method that uses the latent space to estimate the frequency, phase, decay time, and amplitude of each component in noisy multi-component damped sinusoidal signals.
We investigate multi-component cases under Gaussian-distribution training and further examine the effect of the training-data distribution through comparisons between Gaussian and uniform training.
The performance is evaluated through waveform reconstruction and parameter-estimation accuracy.
We find that the proposed method can estimate the parameters with high accuracy even in challenging setups, such as those involving a subdominant component or nearly opposite-phase components, while remaining reasonably robust when the training distribution is less informative. 
This demonstrates its potential as a tool for analyzing short-duration, noisy signals.
\end{abstract}

\section{Introduction}\label{sec:intro}
Exponentially damped sinusoidal signals are widely observed in many physical systems and often play a fundamental role in their dynamical response.
They appear in a broad range of fields, including speech and audio analysis~\cite{McAulay1986}, nuclear magnetic resonance~\cite{ref:NMR}, free-induction-decay optical magnetometry~\cite{ref:FID1,ref:FID2}, and cavity ring-down polarimetry/ellipsometry~\cite{ref:CRDP1,ref:CRDE1}, as well as structural health monitoring~\cite{ref:structure_health_monitoring}, vibration analysis~\cite{ref:vibration_analysis}, radar~\cite{ref:radar}, sonar~\cite{ref:sonar}, communication channels~\cite{ref:communication}, nuclear collective excitations~\cite{ref:nuclear_dipole_review}, and black-hole ringdown in gravitational-wave astronomy~\cite{Berti:2025hly}. 
Depending on the physical setting, such signals may decay rapidly, be contaminated by observational noise, or consist of multiple superposed damped sinusoidal components. 
Their analysis therefore provides access to underlying physical properties across a wide variety of systems.

A broad range of conventional approaches have been developed for this purpose, including least-squares fitting, Fourier-transform-based methods, classical parametric estimators, and subspace-based methods; their properties and limitations have been extensively studied in the signal-processing literature~\cite{ref:Kay1988,ref:StoicaMoses2005,Visschers:21,McAulay1986,Serra1990,Stylianou2001,Jensen2014EDSM,Kafentzis2012eaQHM}. 
In the context of black-hole ringdown, Kumaresan--Tufts, matrix-pencil, and related Prony-type methods have been applied to noisy test signals and numerical-relativity waveforms~\cite{Berti:2007dg,Berti:2007fi}. 
Least-squares-based ringdown fitting has also been refined through iterative extraction procedures~\cite{Takahashi:2023tkb}, in which longer-lived modes are fitted and subtracted sequentially to improve the stability of extraction of subdominant components. 
More recently, Hankel low-rank approaches, including ESPRIT, Cadzow iterations, and IRLS, have been benchmarked for gravitational-wave signal extraction and quasinormal-mode recovery~\cite{Geissler:2026iri}. 
These studies provide important standard tools and baselines for damped-sinusoid parameter estimation, while also showing that the performance can depend sensitively on noise level, model order, signal overlap, and the choice of analysis settings. 
These considerations motivate complementary data-driven approaches that extract compact representations directly from noisy waveform data.

Machine-learning-based methods are promising in such situations because they can learn effective low-dimensional representations of complex signals directly from training data. 
Among them, autoencoders (e.g., \cite{ref:Autoencoder_intro,ref:Autoencoder_rev}) are particularly attractive because they compress high-dimensional input data into a low-dimensional latent space while preserving essential information. 
Such latent representations can capture the dominant structure of noisy signals, thereby reducing sensitivity to noise and supporting both denoising and parameter estimation. 
This makes autoencoders well suited to the present problem, where robust extraction of the physical parameters from noisy multi-component signals is required.

Visschers' group~\cite{ref:autoencoder} demonstrated that an autoencoder can be used for rapid parameter extraction from single-component damped sinusoidal signals\footnote{Autoencoders are usually unsupervised or self-supervised ML models. The framework might be closer to a supervised regression network than to an autoencoder.}, with performance comparable to or better than conventional methods such as least squares~\cite{ref:Halmer} and fast-Fourier-transform-based approaches~\cite{Visschers:21,ref:Bostrom}.
The same study also highlighted the usefulness of training the latent-space representation to encode the physical parameters of interest directly. 
Nevertheless, parameter estimation for noisy multi-component damped sinusoidal signals remains insufficiently explored, especially in difficult situations where one component is subdominant or where partial cancellation occurs between components.

In this paper, building on the single-component study of~\cite{ref:autoencoder}, we developed an autoencoder-based method for estimating the frequency, phase, decay time, and amplitude of each component in noisy multi-component damped sinusoidal signals. 
The aim of the present paper is to provide an initial assessment of the method under controlled synthetic settings and to identify several limitations that should be addressed before applications to more realistic data.
We evaluate the method from two complementary viewpoints: the accuracy of parameter estimation and the reconstruction performance of denoised waveforms, quantified by the match score. 
A preliminary version of the two-component analysis, including a rapidly decaying, low-amplitude component with Gaussian training (\textbf{2G-sub}) and nearly opposite-phase components with Gaussian training (\textbf{2G-opp}) in Table~\ref{tab:all-cases}, was previously reported in a Japanese conference proceedings paper~\cite{ref:Iida2026}. 

The matrix pencil (MP) method is a well-established parametric estimation technique for exponentially damped sinusoids and has been widely used owing to its robustness
and computational efficiency~\cite{ref:MP,ref:MP2}. In the present study, we therefore use a simple MP method as a baseline for comparison in \textbf{2G-sub} and \textbf{2G-opp} cases.
Moreover, the present paper substantially extends that work 
In particular, we show that the autoencoder remains effective even for a five-component superposed damped sinusoidal signal. 
Furthermore, we consider a more practical training setting in which the encoder and decoder are trained using uniform parameter distributions rather than Gaussian ones. 
We then perform a systematic comparison between Gaussian and uniform training distributions in order to examine the effect of the training-data distribution on parameter estimation.
In this way, we comprehensively assess the ability of the proposed autoencoder to estimate the physical parameters of superposed damped sinusoidal signals.

The main contributions of this paper are as follows:
\begin{itemize}
  \item Building on the previous studies~\cite{ref:autoencoder,ref:Iida2026}, we developed an autoencoder-based framework for estimating the physical parameters (frequency, phase, decay time, and amplitude) of superposed multi-component damped sinusoidal signals.
  \item Designed as a baseline study under controlled synthetic settings, with the goal of assessing the basic performance of the method and identifying several observed limitations.
  \item Demonstrated accurate parameter estimation and waveform denoising even under challenging conditions, including subdominant components, partial cancellation between nearly opposite-phase components.
  \item Compared the proposed method with a conventional MP method~\cite{ref:MP,ref:MP2} for representative two-component signal cases, providing a quantitative benchmark against a widely used parametric signal-processing technique.
  \item Systematically investigated the impact of training-data distributions by comparing Gaussian and uniform training, showing that the proposed method remains reasonably robust when prior knowledge of parameter distributions is limited. 
\end{itemize}

The remainder of this paper is organized as follows. 
Section~\ref{sec:method} describes the autoencoder architecture, the data-generation procedure, and the evaluation methods. Section~\ref{sec:results} presents the results and discussion. Finally, Section~\ref{sec:summary} summarizes our findings.

\section{Autoencoder, Data Generation and Evaluation Methods}\label{sec:method}
\begin{figure*}[tbp]
  \centering
  \includegraphics[width=0.95\textwidth]{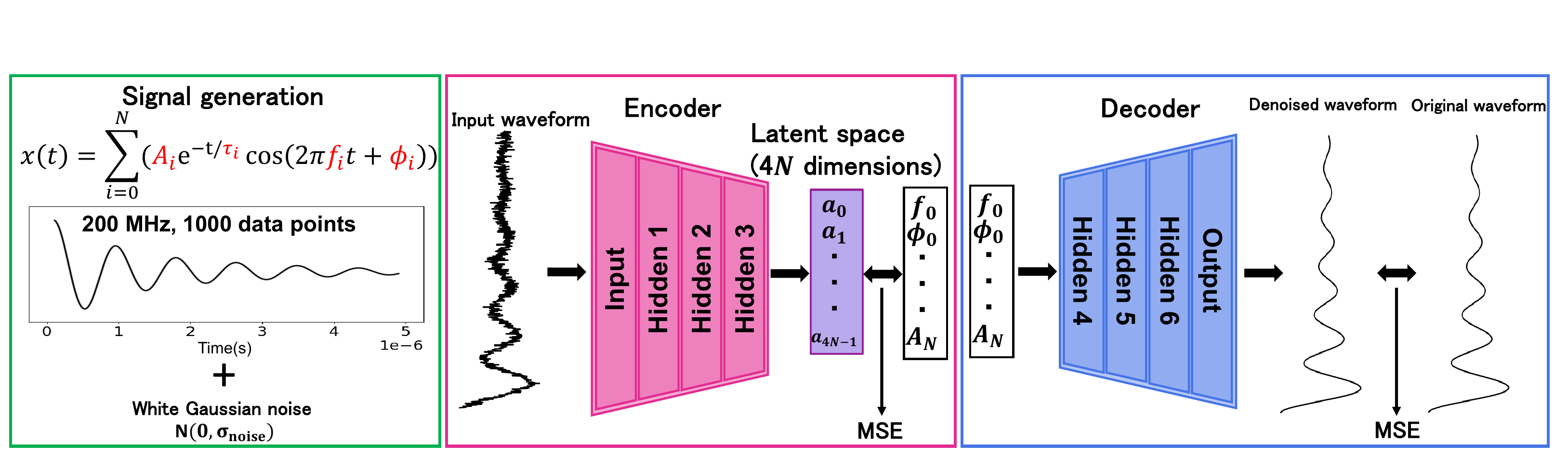}
  \caption{Schematic overview of the analysis pipeline, including data generation (left) and the autoencoder architecture (middle and right). For illustration, the figure shows the two-component case.}
  \label{fig:autoencoder}
\vspace{2mm}
\captionof{table}{
Autoencoder architectures used in this study. The numbers indicate the number of neurons in each layer for each case. The meaning of the compact descriptive case labels is shown in Table~\ref{tab:all-cases}.}
\label{tab:autoencoder_architecture}
\scalebox{0.8}{
\begin{tabular}{cccccc}
\hline\hline
Layer & \textbf{2G-sub}, \textbf{2G-opp} & \textbf{5G} & \textbf{1G}, \textbf{1U} & \textbf{2G}, \textbf{2U} & \textbf{3U} \\
\hline
Input       & 1,000 & 1,000 & 1,000 & 1,000 & 1,000 \\
Hidden  1   & 200   & 500 & 200 & 200 & 200 \\
Hidden  2   & 100   & 200 & 100 & 100 & 100 \\
Hidden  3   & 20    & 50 & 20 & 20 & 20 \\
Latent space  & 8   & 20 & 4 & 8 & 12 \\
Hidden  4   & 20    & 50 & 20 & 20 & 20 \\
Hidden  5   & 100   & 200 & 100 & 100 & 100 \\
Hidden  6   & 200   & 500 & 200 & 200 & 200 \\
Output      & 1,000  & 1,000 & 1,000 & 1,000 & 1,000 \\
\hline\hline
\end{tabular}
}
\end{figure*}

We consider parameter estimation for noisy waveforms composed of superposed multi-component damped sinusoidal signals. 
The analysis is carried out for both two-component and five-component signals to examine how accurately the physical parameters of each component can be extracted. 
This section describes the analysis pipeline, including the autoencoder architecture, the data generation procedure, and the evaluation methods. An overview of the workflow is shown in Fig.~\ref{fig:autoencoder}.

\begin{landscape}
\begin{table}[tbp]
\captionof{table}{Summary of the cases considered in this study. In the Training and Validation columns, G and U denote Gaussian and uniform parameter distributions, respectively. Numbers in parentheses indicate the numbers of training and validation samples used in each case. The compact descriptive case labels encode the number of components, the training distribution (G or U), and, where applicable, the special setting in the suffix.}
\label{tab:all-cases}
\scriptsize
\begin{tabular}{cccccccc}
\hline\hline
Case & Explanation & \#Components & Training & Validation & Noise level~(SNR [dB]) & Result & Special setting \\
\hline
\textbf{2G-sub} & \ref{subsubsec:t_two_G}, Fig.~\ref{fig:case1_parameter-distribution}, Table~\ref{tab:case1_parameter-table} & 2 & G (8,000) & G (2,000) & $1/2~(8.8)$ & \ref{subsubsec:r_two_G}, Fig.~\ref{fig:case1_result}, Table~\ref{tab:case1_table} & Subdominant component \\
\textbf{2G-opp} & \ref{subsubsec:t_two_G}, Fig.~\ref{fig:case2_parameter-distribution}, Table~\ref{tab:case2_parameter-table} & 2 & G (8,000) & G (2,000) & $1/2^3~(14.3)$ & \ref{subsubsec:r_two_G}, Fig.~\ref{fig:case2_result}, Table~\ref{tab:case1_table} & Nearly opposite phases \\
\textbf{5G}  & \ref{subsubsec:t_five_G}, Fig.~\ref{fig:case3_parameter-distribution}, Table~\ref{tab:case3_parameter-table}  & 5 & G (8,000) & G (2,000) & $5~(-5.2)$& \ref{subsubsec:r_five_G}, Figs.~\ref{fig:case3_result}~\ref{fig:case3_result_scatter}, Table~\ref{tab:case3_table}& \\
\textbf{1G} & \ref{subsubsec:t_single_G_U}, Fig.~\ref{fig:case4-5_parameter-distribution}, Table~\ref{tab:case4-5_parameter-table} & 1 & G (4,000) & G (1,000) & $1/2^3~(4.6)$ & \ref{subsubsec:r_single_G_U}, Figs.~\ref{fig:case4_result},~\ref{fig:case4-5_match}, Table~\ref{tab:case4-5_table} & \\
\textbf{1U} & \ref{subsubsec:t_single_G_U}, Fig.~\ref{fig:case4-5_parameter-distribution}, Table~\ref{tab:case4-5_parameter-table}& 1 & U (4,000) & G (1,000) & $1/2^3~(4.6)$  & \ref{subsubsec:r_single_G_U}, Figs.~\ref{fig:case5_result},~\ref{fig:case4-5_match}, Table~\ref{tab:case4-5_table} &\\
\textbf{2G} & \ref{subsubsec:t_two_G_U}, Fig.~\ref{fig:case6-7_parameter-distribution}, Table~\ref{tab:case6-7_parameter-table} & 2 & G (990,000) & G (10,000) & $1/2^3~(15.7)$ & \ref{subsubsec:Two-component}, Fig.~\ref{fig:case6_result}, Table~\ref{tab:case6-7_table} &\\
\textbf{2U} & \ref{subsubsec:t_two_G_U}, Fig.~\ref{fig:case6-7_parameter-distribution}, Table~\ref{tab:case6-7_parameter-table} & 2 & U (1,000,000) & G (10,000) & $1/2^3~(15.7)$ & \ref{subsubsec:Two-component}, Fig.~\ref{fig:case7_result}, Table~\ref{tab:case6-7_table} &\\
\textbf{3U} & \ref{subsubsec:t_three_G_U}, Fig.~\ref{fig:case8_parameter-distribution}, Table~\ref{tab:case8_parameter-table} & 3 & U (1,000,000) & G (10,000) & $1/2^2~(11.4)$ & \ref{subsubsec:Three-component}, Fig.~\ref{fig:case8_result}, Table~\ref{tab:case8_table} &\\
\hline\hline
\end{tabular}
\end{table}
\end{landscape}

\subsection{Overview of autoencoder}\label{subsec:autoencoder}
Figure~\ref{fig:autoencoder} shows the data generation (left) and the autoencoder architecture (middle and right) adopted in this study. 
In this study, we used a nine-layer autoencoder including the input and output layers, and applied the $\tanh$ activation function to the output of each layer except for the latent-space layer. 
No activation function was applied to the latent-space layer so that the latent variables could directly represent the physical parameters without an additional nonlinear restriction.
The network architecture was adjusted according to the number of damped sinusoidal components and the number of parameters to be estimated. 
In particular, the dimension of the latent space was chosen to match the total number of physical parameters to be estimated.
Since each damped sinusoidal component is characterized by four parameters, the dimension of the latent space is equal to four times the number of components.
In all cases, the numbers of neurons were chosen so as to maintain an hourglass-shaped structure around the latent-space layer.
The detailed network architectures for each case are summarized in Table~\ref{tab:autoencoder_architecture}, while the corresponding data-generation settings are described in Section~\ref{subsec:data_generation} (see Table~\ref{tab:all-cases}). 

First, the encoder takes a noisy waveform composed of superposed damped sinusoidal components as input and maps it to the latent space. 
The input waveforms for the encoder were standardized in order to stabilize the training. 
In general, the latent space of an autoencoder does not have to coincide directly with the physical parameters of interest. 
In the present study, such a correspondence was enforced through the training procedure of the encoder. 
Specifically, following~\cite{Visschers:21}, we set the dimension of the latent space to match the total number of physical parameters. 

For example, in the two-component cases, the latent space has eight dimensions corresponding to the four parameters (frequency, phase, decay time, and amplitude) for each of the two components. 
We then adopted the mean squared error (MSE) between the estimated parameters and the true parameters as the loss function for encoder training, so that the latent variables correspond to the physical parameters of each damped sinusoidal component.
In the previous study~\cite{Visschers:21}, stochastic gradient descent (SGD)~\cite{ref:SGD} was adopted for optimization. 
In the present study, by contrast, we employed Adam~\cite{ref:adam}, because it led to faster convergence and reduced computational time.

The physical parameters (frequency, phase, decay time, or amplitude) were normalized before being provided to the decoder.
This normalization was introduced to stabilize the training.
For both Gaussian and uniform training distributions, we used
\begin{equation}
p_{\mathrm{lat}} = \frac{p-\mu_p}{3\sigma_p},
\label{eq:ls_norm}
\end{equation}
where $\mu_p$ and $\sigma_p$ denote the mean and standard deviation of the corresponding training-data distribution for the parameter $p$, respectively.
The factor of $3$ was introduced so that most samples fall within a range of order unity, which was found to be suitable for stable training~\cite{Visschers:21}.
Although we follow~\cite{ref:autoencoder} and refer to this framework as an autoencoder, the encoder is trained using labeled parameter values and therefore contains a supervised component.

Next, the decoder was trained to reconstruct the corresponding denoised waveform from the set of physical parameters represented in the latent space. 
Specifically, the four parameters of each damped sinusoidal component were used as inputs to the decoder. 
For decoder training, the MSE between the denoised waveform and the original noise-free waveform was used as the loss function, and the optimization was again performed using Adam. 
In this way, the encoder and decoder were trained separately, adopting a framework in which the physical parameters can be estimated in the latent space.

Moreover, to mitigate overfitting, dropout with a rate of 0.1 was applied to each layer of both the encoder and the decoder in all cases.
In this procedure, a fraction of neurons is randomly deactivated during training, which prevents excessive co-adaptation of the network and improves generalization performance.

\subsection{Data generation}\label{subsec:data_generation}
The time-series waveform of a superposition of damped sinusoidal components is written as
\begin{equation}
x(t) = \sum_{i=0}^{N} A_i  e^{-t/\tau_i}  \cos(2\pi f_i t + \phi_i),
\label{eq:multi_damped}
\end{equation}
where the damped sinusoidal components are indexed by $i=0,1,\dots,N$, so that the total number of components is $N+1$. 
Each damped sinusoidal component is characterized by the parameters $A_i$, $\tau_i$, $f_i$, and $\phi_i$, which represent the amplitude, decay time, frequency, and initial phase of the $i$th component, respectively. 
$t$ denotes the elapsed time. 
Following Visschers' group~\cite{ref:autoencoder}, we set the signal length to 5~$\mu$s and the sampling rate to 200~MHz, resulting in time-series data with 1,000 sample points (Fig.~\ref{fig:autoencoder}, left).

To construct the dataset, we generated a large number of superposed damped sinusoidal waveforms by varying the physical parameters of each component over prescribed distributions. 
For each damped sinusoidal component, the parameters $f_i$, $\phi_i$, $\tau_i$, and $A_i$ were drawn from Gaussian distributions with mean $\mu_{\rm{par}}$ and standard deviation $\sigma_{\rm{par}}$.
The original noise-free waveform in Eq.~(\ref{eq:multi_damped}) was constructed from these parameter values, and the input waveform was generated by adding white Gaussian noise with mean $0$ and standard deviation $\sigma_{\rm{noise}}$.
We performed the analysis for several noise levels.
Among these, one representative result for each case is presented and discussed in Section~\ref{sec:results}.

In the following, we consider a range of cases to comprehensively assess the ability of the proposed autoencoder to estimate the physical parameters of superposed damped sinusoidal signals. 
These cases were designed to probe different aspects of the problem, including variations in the number of superposed components, the choice of training-data distribution, and deliberately challenging setups such as subdominant components and nearly opposite-phase superpositions. 
The cases considered in this study are summarized in Table~\ref{tab:all-cases}.
The compact descriptive case labels encode the number of components, the training distribution (G or U), and, where applicable, the special setting in the suffix.

For \textbf{2G-sub}, \textbf{2G-opp} and \textbf{5G}, both the training and validation data were generated from Gaussian parameter distributions, providing a controlled setting for performance evaluation. 
For each value of $\sigma_{\rm{noise}}$, a total of 10,000 samples were generated, of which 20\% were used for validation.

In actual parameter-estimation problems, however, the true parameter values are not known a priori, and it is often more natural to assume only a broad parameter range rather than a specific Gaussian prior. 
Training with uniform parameter distributions therefore provides a useful test of whether the method still performs reasonably when the prior knowledge of the parameter distributions is less informative. 
We examine such uniform-training cases and compare them with the corresponding Gaussian-training cases. 
The numbers of training and validation samples, as well as the noise levels, are summarized in Table~\ref{tab:all-cases}.
The number of training samples was increased for the uniform-training cases and for cases with larger parameter spaces in order to sample the broader distributions more densely. The validation set was generated independently from the training set and used only for performance evaluation.

For all cases, the training was completed within a few hours on a workstation equipped with an Intel(R) Xeon(R) CPU E5-2637 v4 @ 3.50GHz, an NVIDIA GeForce RTX 2080 Ti GPU, and 128 GB memory.

\begin{figure*}[tbp]
  \centering
   \begin{subfigure}{0.6\linewidth}
   \includegraphics[width=80mm]{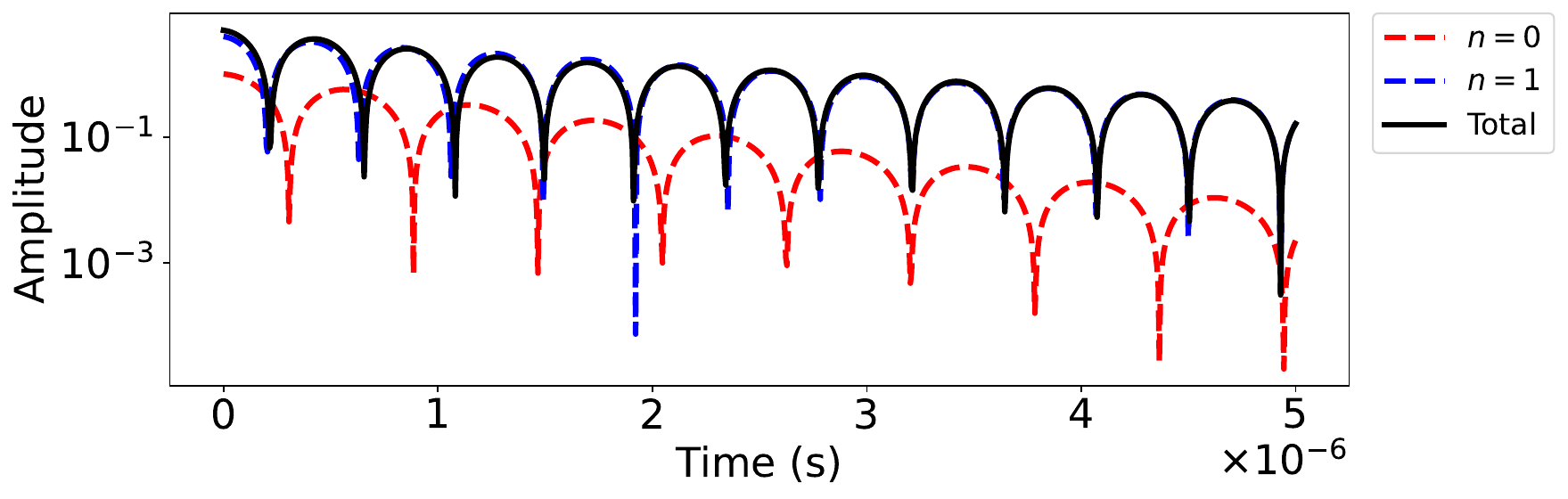}
   \caption{Example of the absolute value of the waveform plotted on a logarithmic scale.}
   \end{subfigure}
   \begin{subfigure}{0.6\linewidth}
  \includegraphics[width=80mm]{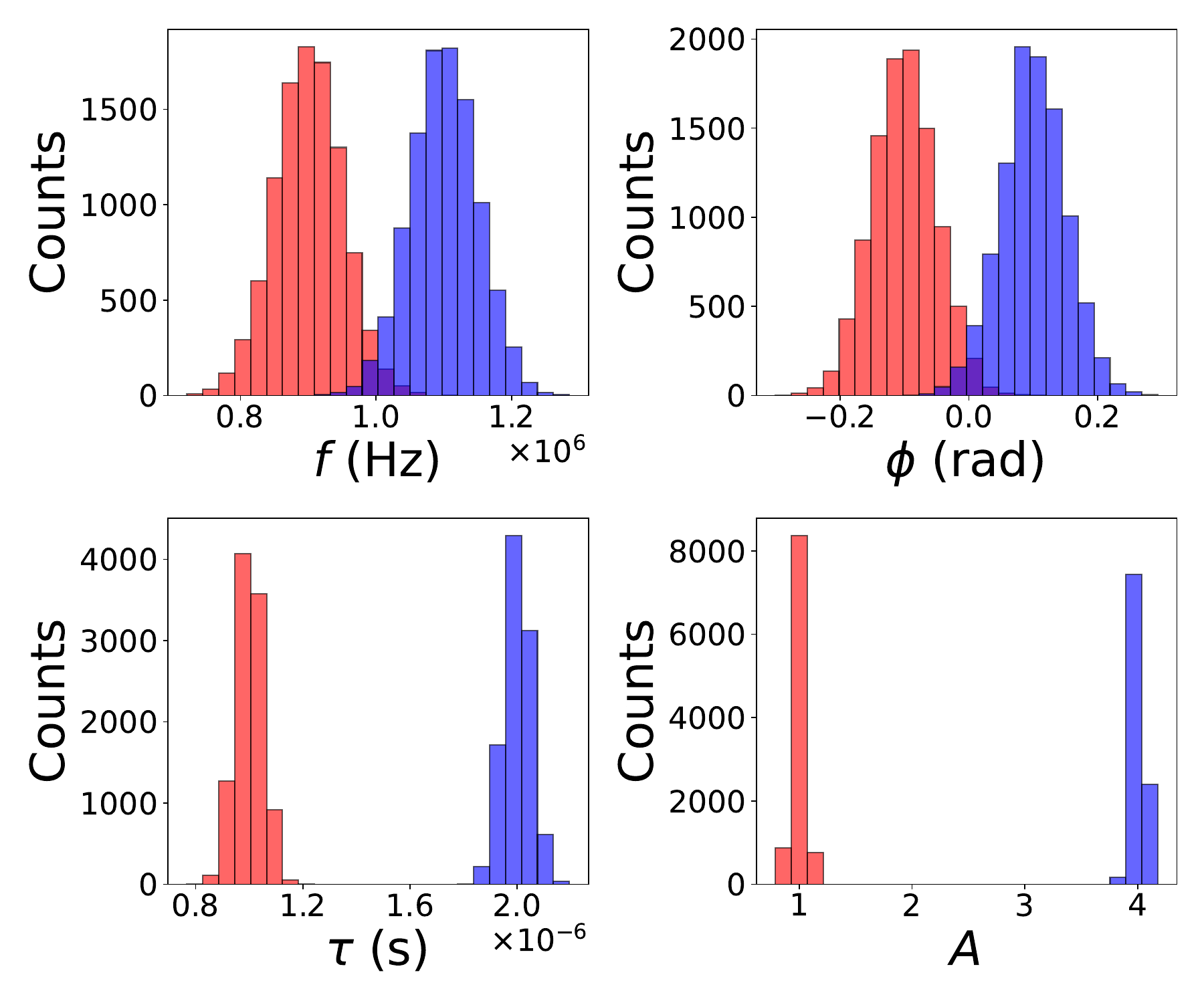}
   \caption{Distributions of the generated true parameters.}
   \end{subfigure}
  \caption{\textbf{Case 1}: A two-component waveform including a rapidly decaying, low-amplitude component.}
  \label{fig:case1_parameter-distribution}
\vspace{2mm}
\captionof{table}{Mean values $\mu_{\rm par}$ and standard deviations $\sigma_{\rm par}$ of the Gaussian distributions used to generate the true parameters in \textbf{2G-sub}. The resulting parameter distributions are shown in Fig.~\ref{fig:case1_parameter-distribution}(b).} \label{tab:case1_parameter-table}
\scalebox{0.8}{
\begin{tabular}{c c c c}
\hline\hline
Parameter & Component $i$ & $\mu_{\rm{par}}$ & $\sigma_{\rm{par}}$ \\
\hline
$f_{i}$ [MHz] & 0 (red) & $0.90$ & $0.050$ \\
& 1 (blue) & $1.1$ & $0.050$\\
\hdashline
$\phi_{i}$ [rad] & 0 (red) & $-0.10$ & $0.050$ \\
& 1 (blue) & $0.10$ & $0.050$\\
\hdashline
$\tau_{i}$ [$\mu$s] & 0 (red) & $1.0$ & $0.050$ \\
& 1 (blue) & $2.0$ & $0.050$\\
\hdashline
$A_{i}$ & 0 (red) & $1.0$ & $0.050$ \\
& 1 (blue) & $4.0$ & $0.050$\\
\hline\hline
\end{tabular}
}
\end{figure*}
%
\begin{figure*}[tbp]
  \centering
   \begin{subfigure}{0.6\linewidth}
   \includegraphics[width=80mm]{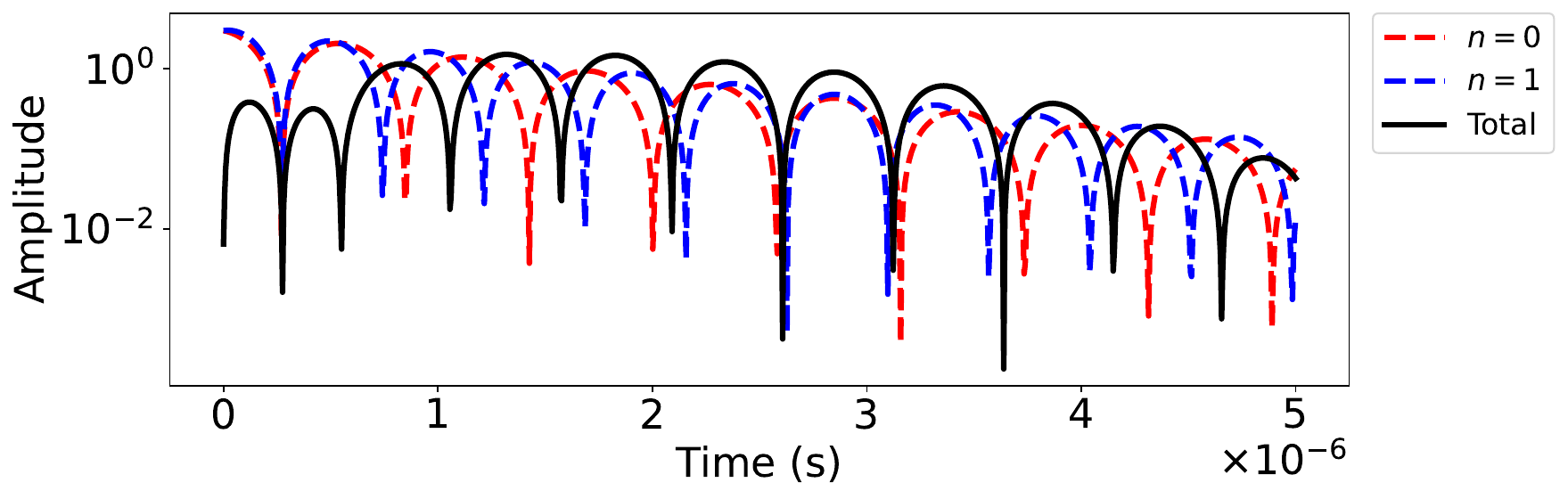}
   \caption{Example of the absolute value of the waveform plotted on a logarithmic scale.}
   \end{subfigure}
    \begin{subfigure}{0.6\linewidth}
  \includegraphics[width=80mm]{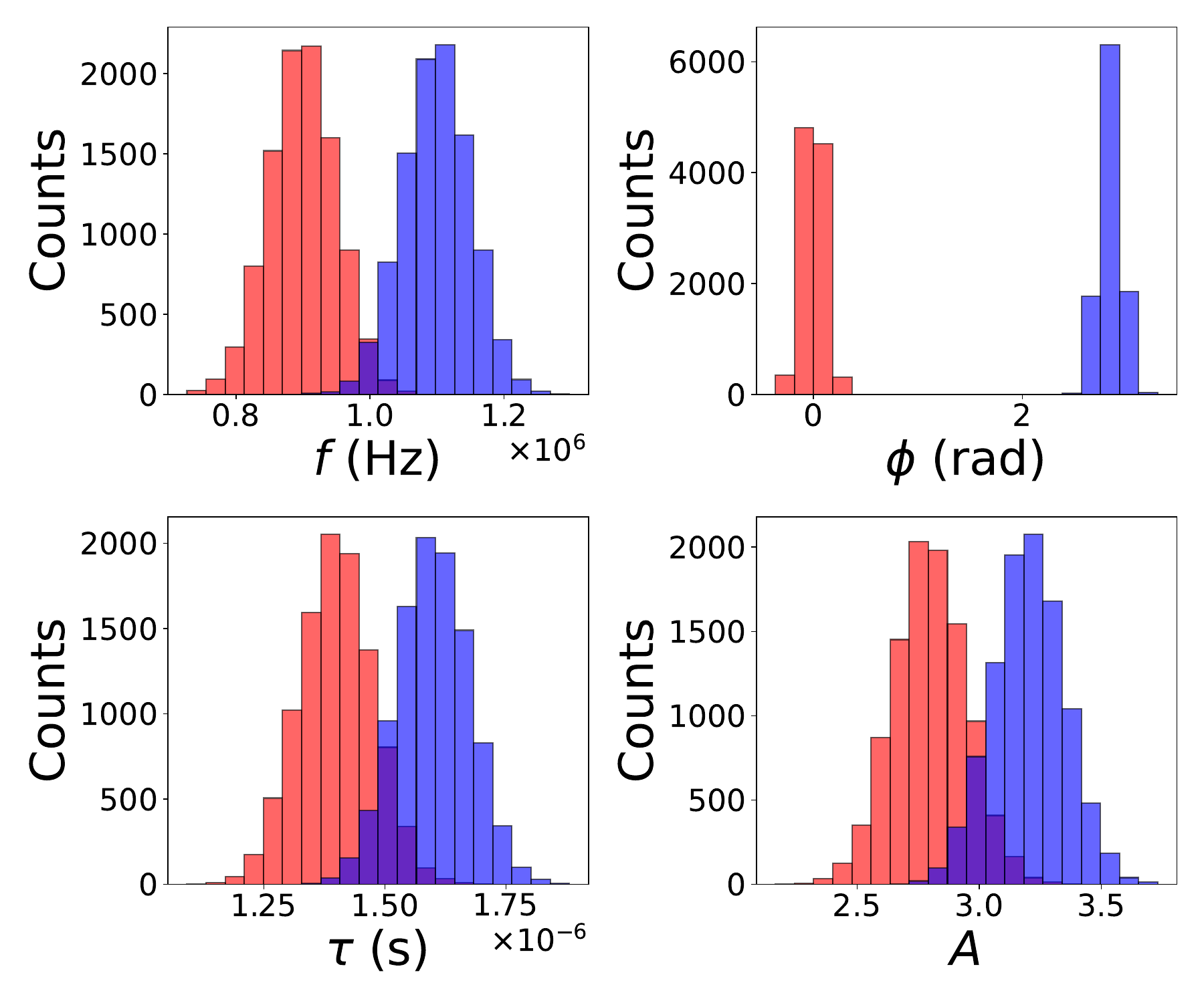}
   \caption{Distributions of the generated true parameters.}
   \end{subfigure}
   \caption{\textbf{2G-opp}: A two-component waveform with nearly opposite phases.}
  \label{fig:case2_parameter-distribution}
\vspace{2mm}
\captionof{table}{Mean values $\mu_{\rm par}$ and standard deviations $\sigma_{\rm par}$ of the Gaussian distributions used to generate the true parameters in \textbf{2G-opp}. The resulting parameter distributions are shown in Fig.~\ref{fig:case2_parameter-distribution}(b).}
\label{tab:case2_parameter-table}
\scalebox{0.8}{
\begin{tabular}{c c c c}
\hline\hline
Parameter & Component $i$ & $\mu_{\rm{par}}$ & $\sigma_{\rm{par}}$ \\
\hline
$f_{i}$ [MHz] & 0 (red) & $0.90$ & $0.050$ \\
& 1 (blue) & $1.1$ & $0.050$\\
\hdashline
$\phi_{i}$ [rad] & 0 (red) & $0.00$ & $0.10$ \\
& 1 (blue) & $2.8$ & $0.10$\\
\hdashline
$\tau_{i}$ [$\mu$s] & 0 (red) & $1.4$ & $0.070$ \\
& 1 (blue) & $1.6$ & $0.070$\\
\hdashline
$A_{i}$ & 0 (red) & $2.8$ & $0.15$ \\
& 1 (blue) & $3.2$ & $0.15$\\
\hline\hline
\end{tabular}
} 
\end{figure*}
%
\begin{figure*}[tbp]
  \centering
  \begin{subfigure}{0.6\linewidth}
  \includegraphics[width=80mm]{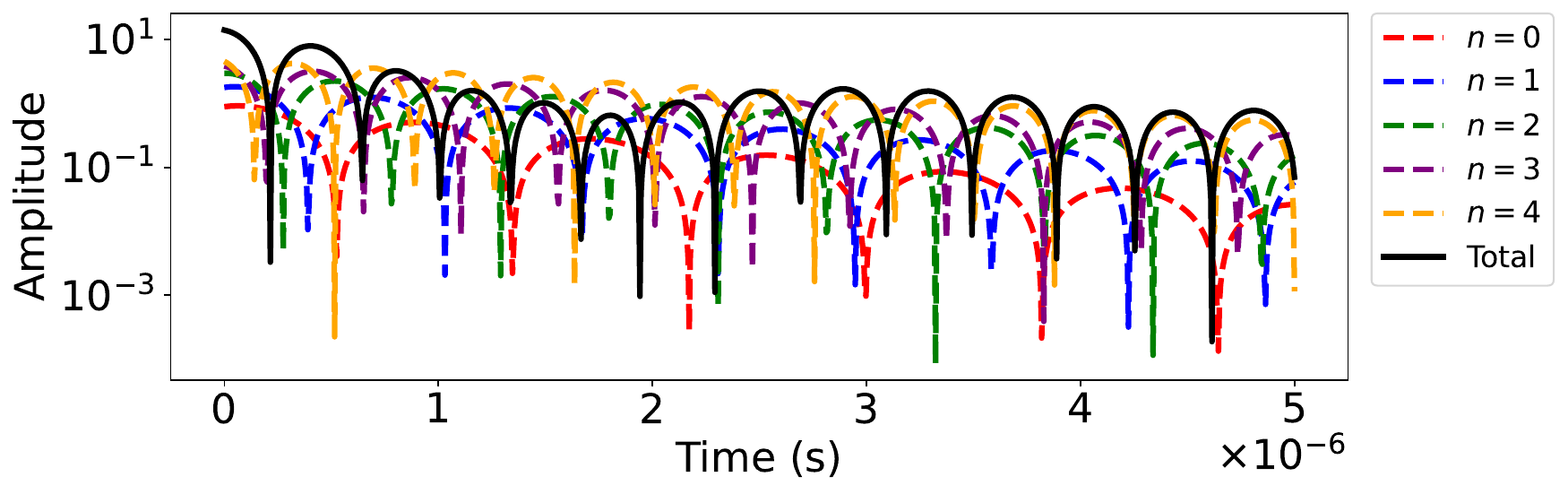}
   \caption{Example of the absolute value of the waveform plotted on a logarithmic scale.}
   \end{subfigure}
   \begin{subfigure}{0.60\linewidth}
  \includegraphics[width=80mm]{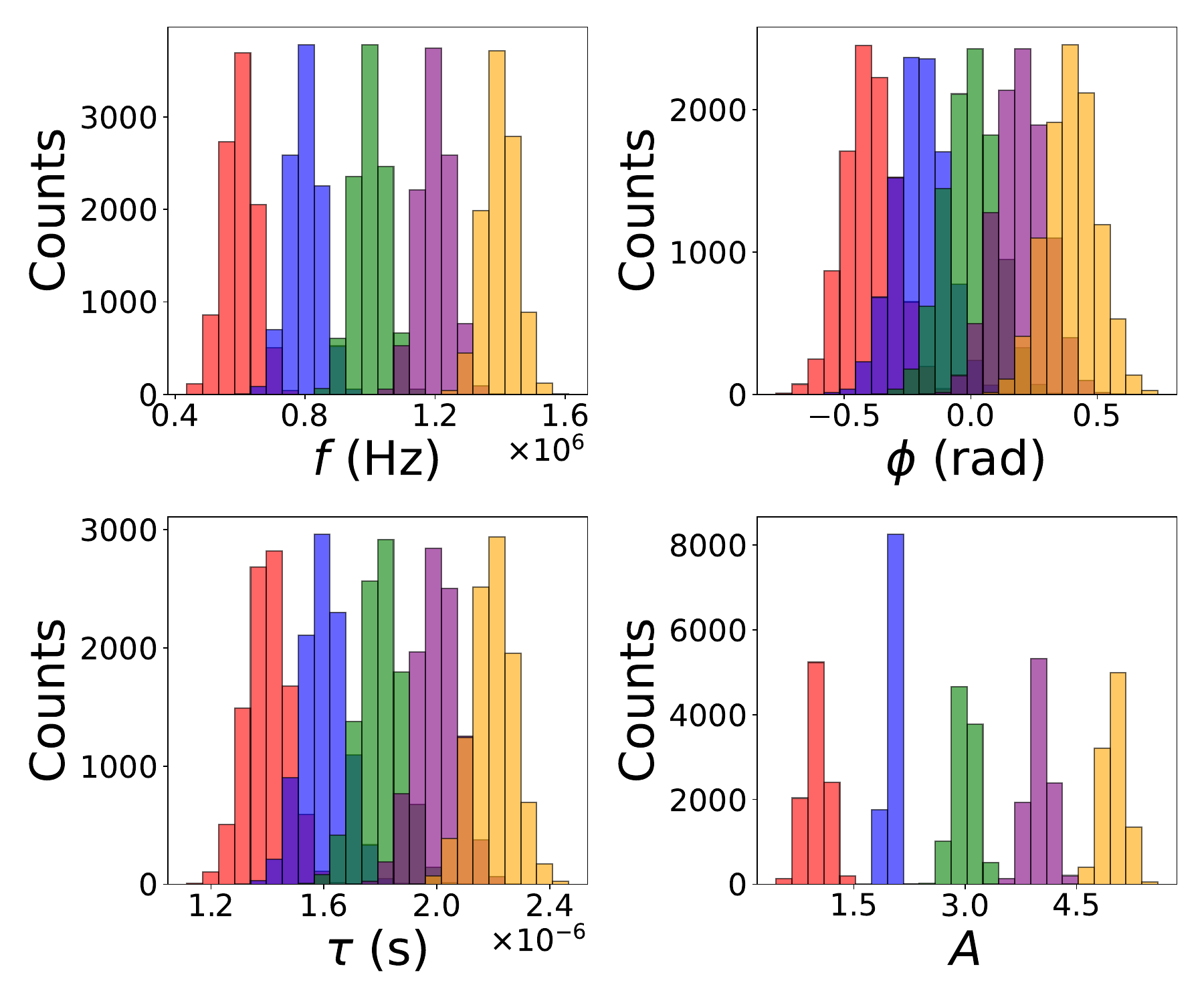}
   \caption{Distributions of the generated true parameters.}
   \end{subfigure}
  \caption{\textbf{5G}: A five-component superposed damped sinusoidal waveform.}
  \label{fig:case3_parameter-distribution}
\vspace{2mm}
\captionof{table}{Mean values $\mu_{\rm par}$ and standard deviations $\sigma_{\rm par}$ of the Gaussian distributions used to generate the true parameters in \textbf{5G}. The resulting parameter distributions are shown in Fig.~\ref{fig:case3_parameter-distribution}(b).
}\label{tab:case3_parameter-table}
\scalebox{0.7}{
\begin{tabular}{c c c c}
\hline\hline
Parameter & Component $i$ & $\mu_{\rm{par}}$ & $\sigma_{\rm{par}}$ \\
\hline
$f_{i}$ [MHz] & 0 (red) & $0.60$ & $0.10$ \\
& 1 (blue) & $0.80$ & $0.10$ \\
& 2 (green) & $1.0$ & $0.10$ \\
& 3 (purple) & $1.2$ & $0.10$ \\
& 4 (yellow) & $1.4$ & $0.10$ \\
\hdashline
$\phi_{i}$ [rad] & 0 (red) & $-0.40$ & $0.10$ \\
& 1 (blue) & $-0.20$ & $0.10$\\
& 2 (green) & $0.00$ & $0.10$ \\
& 3 (purple) & $0.20$ & $0.10$ \\
& 4 (yellow) & $0.40$ & $0.10$ \\
\hdashline
$\tau_{i}$ [$\mu$s] & 0 (red) & $1.4$ & $0.10$ \\
& 1 (blue) & $1.6$ & $0.10$\\
& 2 (green) & $1.8$ & $0.10$ \\
& 3 (purple) & $2.0$ & $0.10$ \\
& 4 (yellow) & $2.2$ & $0.10$ \\
\hdashline
$A_{i}$ & 0 (red) & $1.0$ & $0.10$ \\
& 1 (blue) & $2.0$ & $0.10$\\
& 2 (green) & $3.0$ & $0.10$ \\
& 3 (purple) & $4.0$ & $0.10$ \\
& 4 (yellow) & $5.0$ & $0.10$ \\
\hline\hline
\end{tabular}
} 
\end{figure*}

\subsubsection{Two-component superposed damped sinusoidal signals: Gaussian training}\label{subsubsec:t_two_G}
We first summarize the two-component cases with Gaussian-distribution training data, which were reported in preliminary form in~\cite{ref:Iida2026}, in order to provide a baseline for the more general multi-component analyses presented below.
Among waveforms composed of two superposed damped sinusoidal components, we focus on two cases in which parameter estimation for the individual components is expected to be particularly difficult.

\textbf{2G-sub} is intended to test parameter estimation when one of the two components is subdominant because of its rapid decay and small amplitude, as shown in Fig.~\ref{fig:case1_parameter-distribution}(a).
The parameters of the two components were drawn from Gaussian distributions with mean and standard deviation listed in Table~\ref{tab:case1_parameter-table}, and the resulting parameter distributions are shown in Fig.~\ref{fig:case1_parameter-distribution}(b). 
As in~\cite{ref:autoencoder}, the means of the frequency and phase distributions were shifted to help distinguish between components 0 and 1. 
Component 0 was assigned a shorter decay time and a smaller amplitude, making it more difficult to identify in the superposed signal.
As discussed in Section~\ref{subsec:data_generation}, we examined several noise levels, $\sigma_{\rm{noise}} = 1/2$, $1/2^2$, and $1/2^3$. 
For \textbf{2G-sub}, we focus on the result for $\sigma_{\rm{noise}} = 1/2$, (signal-to-noise ratio, $\rm{SNR} \sim 8.8 \rm{dB}$)\footnote{The signal-to-noise ratio for $k$-th waveform is defined as $\mathrm{SNR}_k~\rm{[dB]}=10\log_{10}\!\left[\sum_{j=0}^{N_t-1}|x_k(t_j)|^2/\sum_{j=0}^{N_t-1}|n_k(t_j)|^2\right]$, where $x_k(t_j)$ is a signal, $n_k(t_j)$ is a noise and $N_{t}$ is the number of time samples ($N_{t}=1000$). The displayed value here is the mean of the finite-sample SNRs of the validation waveforms.}

\textbf{2G-opp} corresponds to the waveform shown in Fig.~\ref{fig:case2_parameter-distribution}(a), in which the two components can partially cancel each other in the superposed waveform. 
The parameters of the two components were drawn from Gaussian distributions with mean and standard deviation listed in Table~\ref{tab:case2_parameter-table}, and the resulting parameter distributions are shown in Fig.~\ref{fig:case2_parameter-distribution}(b). 
The phase distributions were chosen such that the mean phase of component 0 was near $0$, whereas that of component 1 was near $\pi$, so that the two components have nearly opposite phases. 
For \textbf{2G-opp}, we focus on the result for $\sigma_{\rm{noise}} = 1/2^3$ ($\rm{SNR} \sim 14.3 \rm{dB}$), because partial cancellation can reduce the amplitude of the superposed signal and make parameter estimation more difficult.

For \textbf{2G-sub} and \textbf{2G-opp}, a simple MP method~\cite{ref:MP,ref:MP2} was additionally applied to the same validation datasets to provide a baseline comparison with a conventional parametric signal-processing method.

\subsubsection{Five-component superposed damped sinusoidal signals: Gaussian training}\label{subsubsec:t_five_G}
We next considered a more challenging case, \textbf{5G}, consisting of five superposed damped sinusoidal components, as illustrated by the example waveform in Fig.~\ref{fig:case3_parameter-distribution}(a). 
The parameters of the five components were drawn from Gaussian distributions with mean and standard deviation listed in Table~\ref{tab:case3_parameter-table}, and the resulting parameter distributions are shown in Fig.~\ref{fig:case3_parameter-distribution}(b). 
In this case, the amplitudes, decay times, frequencies, and phases were all varied across the five components, increasing the complexity of the signal and making parameter estimation for the individual components more difficult. 

For \textbf{5G}, we focus on the result for $\sigma_{\rm{noise}} = 5$ ($\rm{SNR} \sim -5.2 \rm{dB}$). 
Although the five-component case is more complex in terms of the number of superposed modes, it does not involve the particularly unfavorable configurations considered in \textbf{2G-sub} and \textbf{2G-opp}, such as a strongly subdominant component or partial cancellation between nearly opposite-phase components. 
This allows the parameter estimation to remain effective even at a higher noise level, and we therefore adopt $\sigma_{\rm{noise}} = 5$ as a representative setting for \textbf{5G}.

\subsubsection{Single-component damped sinusoidal signal: Gaussian vs Uniform training}\label{subsubsec:t_single_G_U}

\begin{figure*}[tbp]
  \centering
  \begin{subfigure}{0.3\linewidth}
    \centering
    \includegraphics[width=45mm]{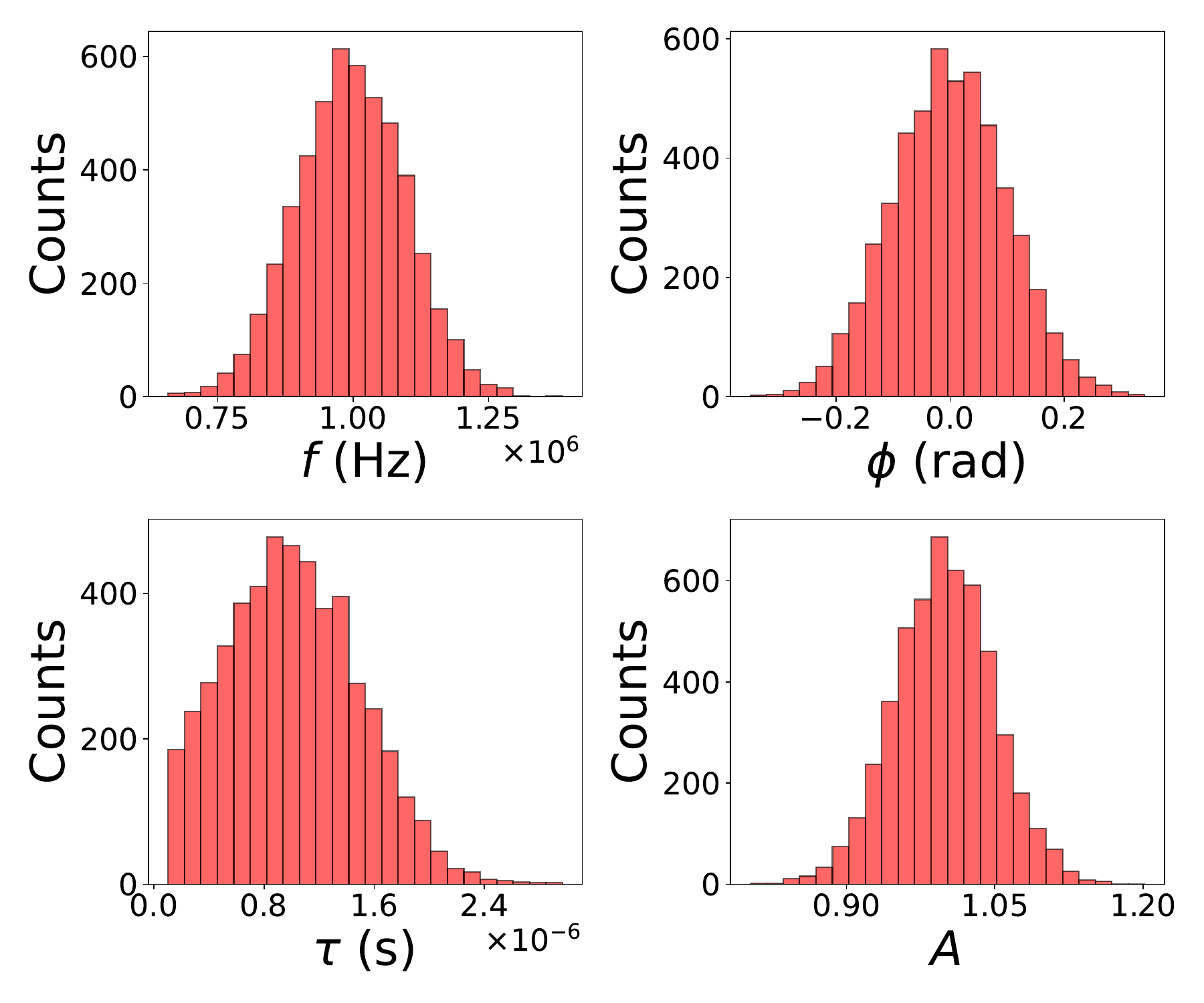} 
    \caption{Gaussian training data.}    
  \end{subfigure}
  \begin{subfigure}{0.3\linewidth}
    \centering
    \includegraphics[width=45mm]{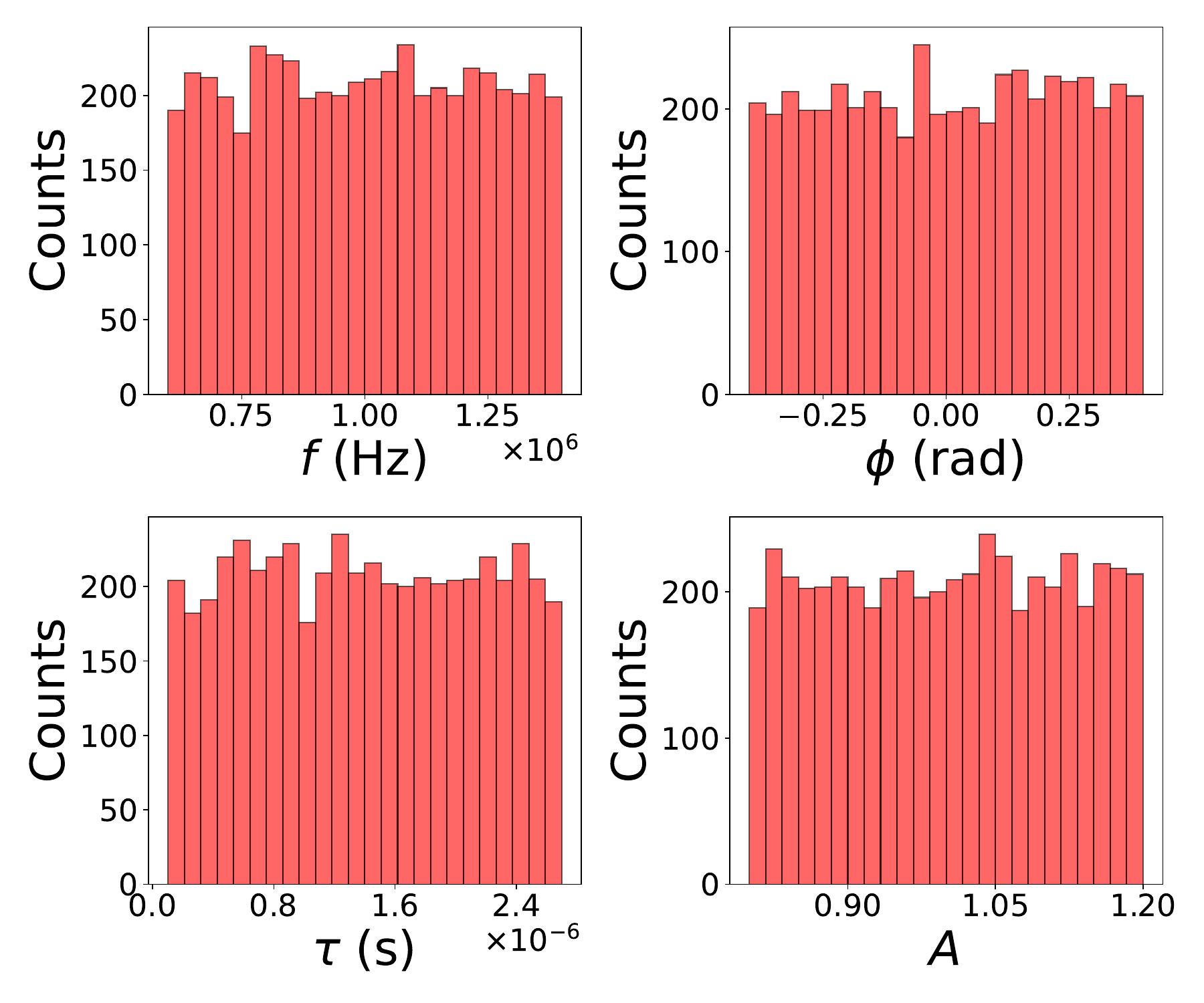}
    \caption{Uniform training data.}
  \end{subfigure}
  \begin{subfigure}{0.3\linewidth}
    \centering
    \includegraphics[width=45mm]{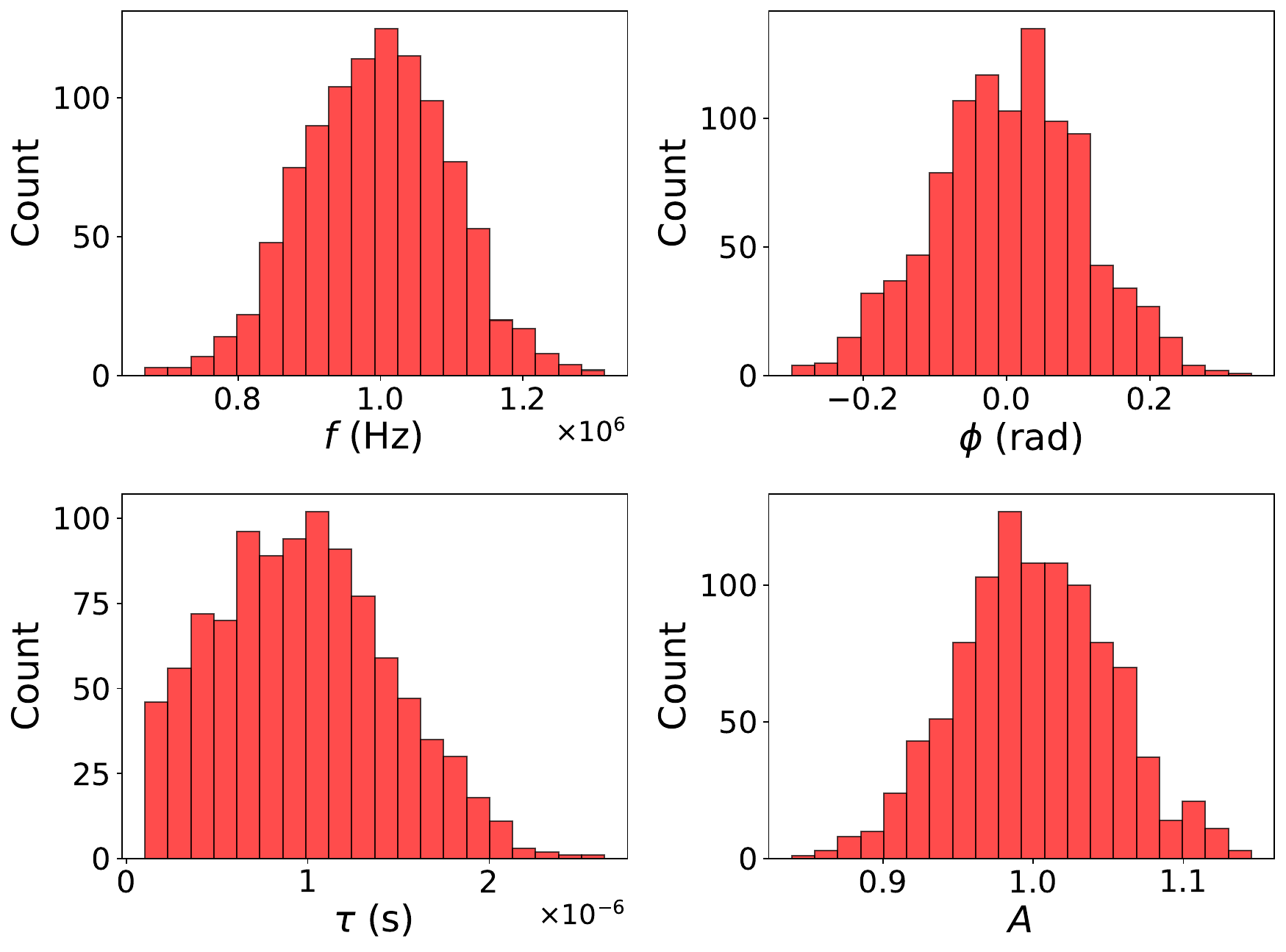}
    \caption{Validation data.}
  \end{subfigure}
  \caption{\textbf{1G} and \textbf{1U}: Training and validation parameter distributions for the single-component case.}
  \label{fig:case4-5_parameter-distribution}
\vspace{2mm}
\captionof{table}{The mean values $\mu_{\rm{par}}$ and standard deviations $\sigma_{\rm{par}}$ of the Gaussian parameter distributions used for the \textbf{1G} training data and the validation data for both \textbf{1G} and \textbf{1U}, and the range of uniform parameter distributions used for the \textbf{1G} training data in the single-component case.}
\label{tab:case4-5_parameter-table}
\scalebox{0.8}{
\begin{tabular}{c c c c}
\hline\hline
\textbf{Gaussian} & $\mu_{\rm{par}}$ & $\sigma_{\rm{par}}$ \\
\hline
$f_{0}$ [MHz] & $1.0$ & $0.10$ \\
\hdashline
$\phi_{0}$ [rad] & $0.00$ & $0.10$ \\
\hdashline
$\tau_{0}$ [$\mu$s] & $1.0$ & $0.50$ \\
\hdashline
$A_{0}$ & $1.0$ & $0.050$ \\
\hline
\textbf{Uniform} & Minimum & Maximum \\
\hline
$f_{0}$ [MHz] & $0.60$ & $1.4$ \\
\hdashline
$\phi_{0}$ [rad] & $-0.40$ & $0.40$ \\
\hdashline
$\tau_{0}$ [$\mu$s] & $0.10$ & $2.7$ \\
\hdashline
$A_{0}$ & $0.80$ & $1.2$ \\
\hline\hline
\end{tabular}
}
\end{figure*}

We next examine the effect of the training-data distribution. 
First, we considered a single-component damped sinusoidal signal with training data generated from Gaussian distributions (\textbf{1G}) and uniform distributions (\textbf{1U}).
The waveforms were generated according to Eq.~(\ref{eq:multi_damped}), and white Gaussian noise with standard deviation $\sigma_{\rm{noise}} = 1/2^3$ ($\rm{SNR} \sim 4.6 \rm{dB}$) was added. 
The training samples were generated from Gaussian and uniform parameter distributions for \textbf{1G} and \textbf{1U}, respectively.
The validation samples were generated from the same Gaussian parameter distributions as the training data for \textbf{1G}.
The range of the uniform distribution in \textbf{1U} was chosen to cover the parameter range of the validation samples. 
The parameter distributions are shown in Fig.~\ref{fig:case4-5_parameter-distribution}, and the parameters used for each distribution are listed in Table~\ref{tab:case4-5_parameter-table}.

For the decay time $\tau$, we followed~\cite{ref:autoencoder} and used the absolute value of the Gaussian random variable, so that only positive decay times were retained. 
This treatment is necessary because negative values of $\tau$ would correspond to exponentially increasing, rather than decaying, signals.

\subsubsection{Two-component superposed damped sinusoidal signals: Gaussian vs Uniform training}\label{subsubsec:t_two_G_U}
\begin{figure*}[tbp]
  \centering
  \begin{subfigure}{0.3\linewidth}
    \centering
    \includegraphics[width=45mm]{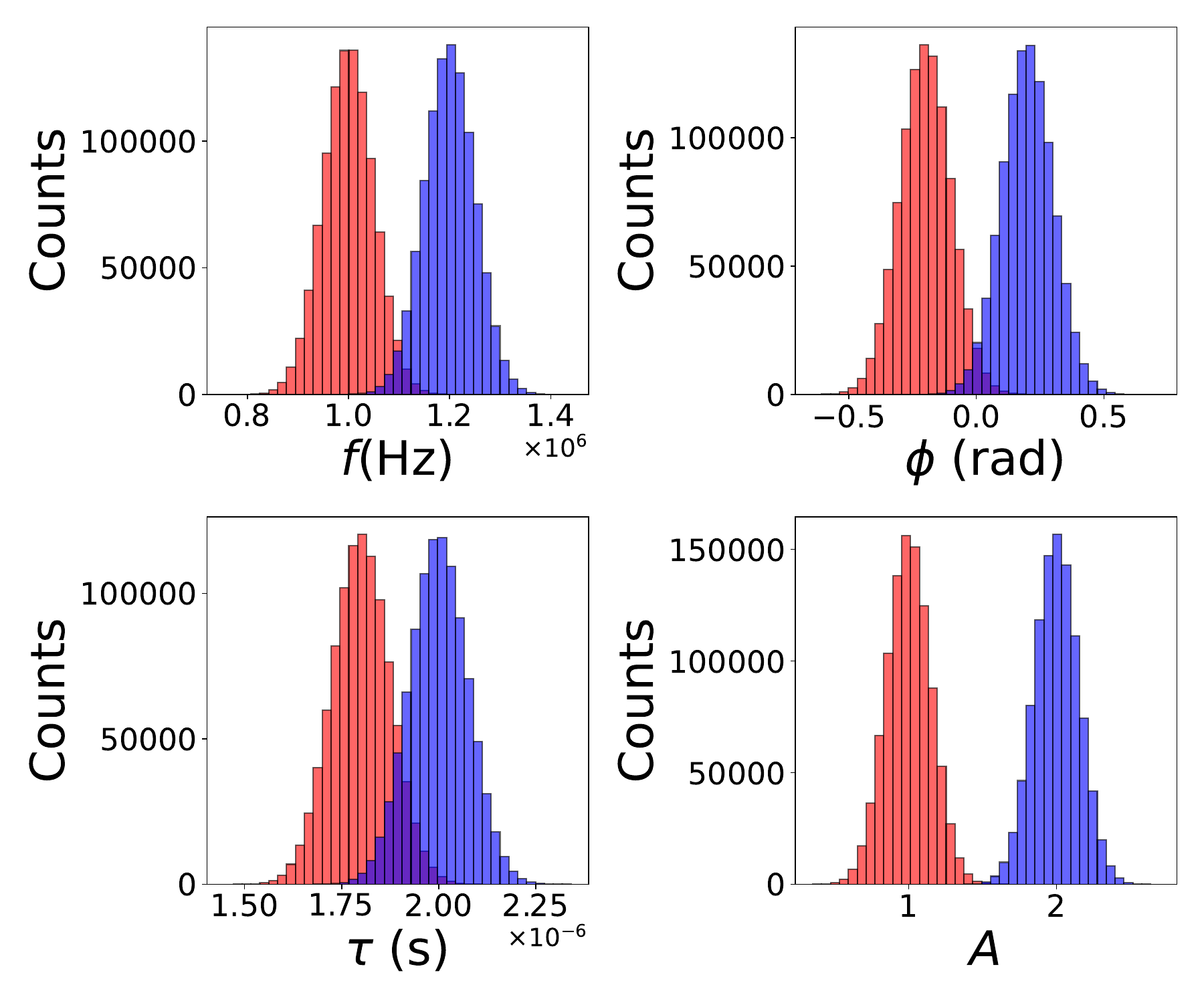} 
    \caption{Gaussian training data.}
  \end{subfigure}
  \begin{subfigure}{0.3\linewidth}
    \centering
    \includegraphics[width=45mm]{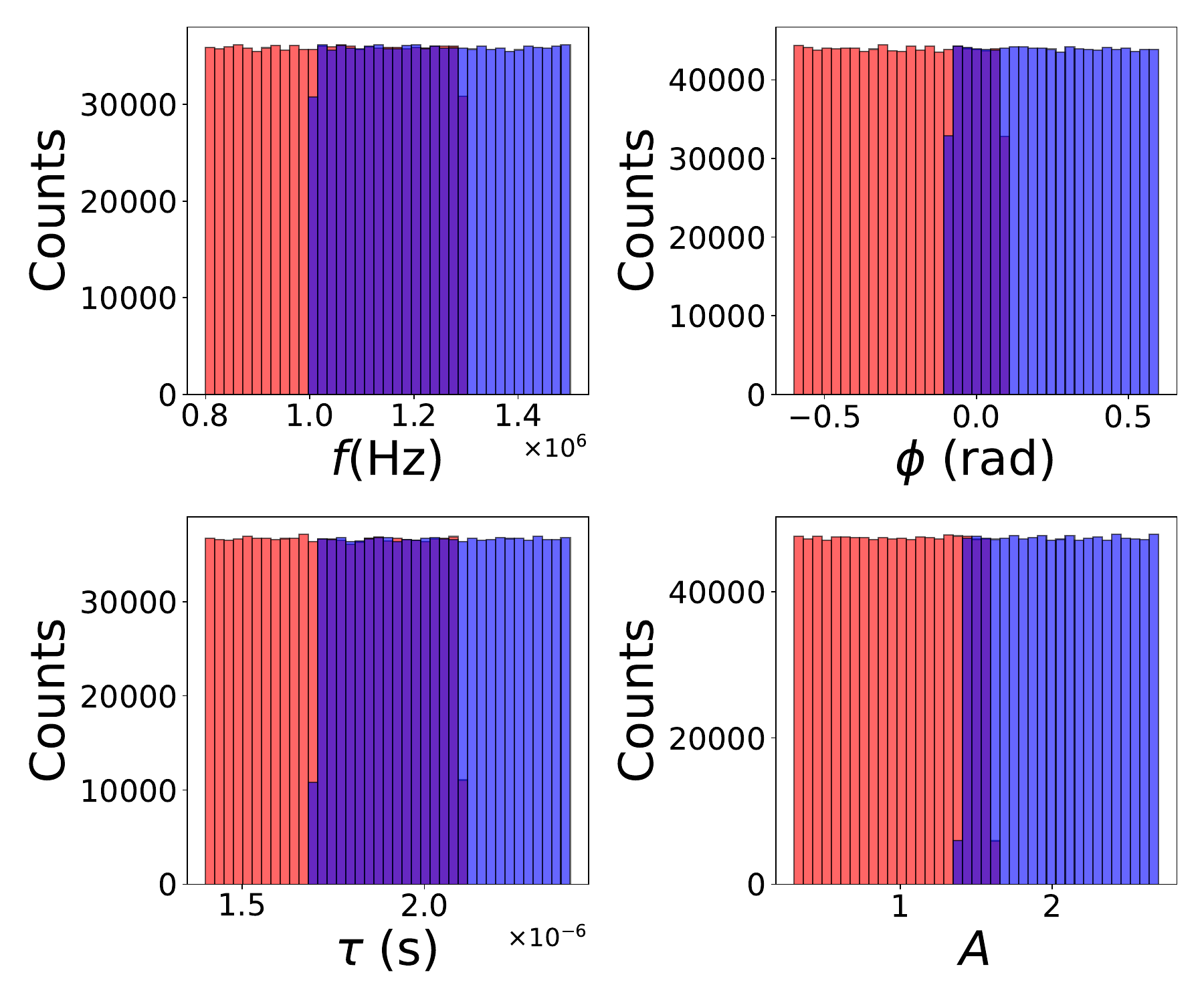}
    \caption{Uniform training data.}
  \end{subfigure}
  \begin{subfigure}{0.3\linewidth}
    \centering
    \includegraphics[width=45mm]{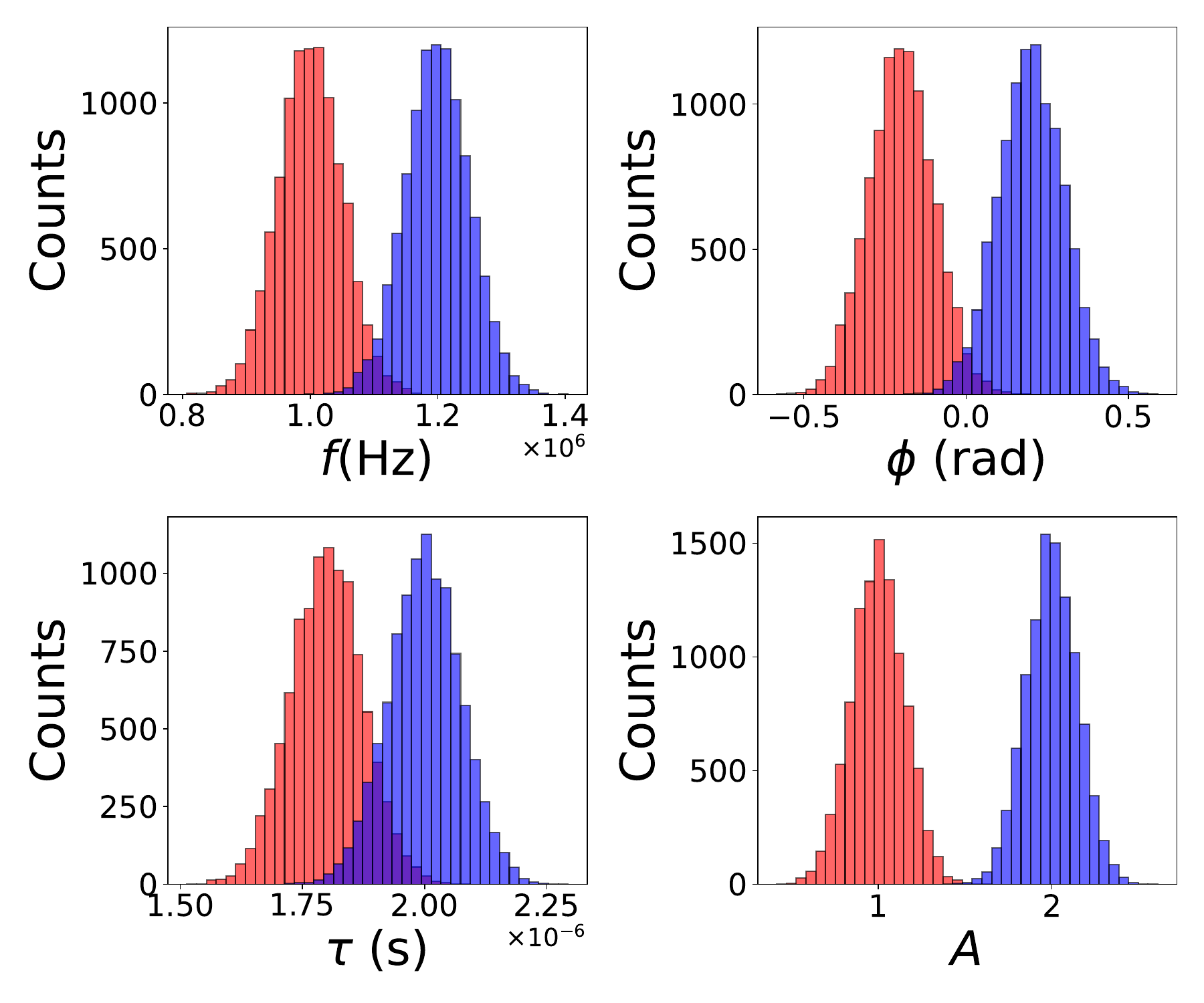}
    \caption{Validation data.}
  \end{subfigure}
  \caption{\textbf{2G} and \textbf{2U}: Training and validation parameter distributions for the two-component case.}
  \label{fig:case6-7_parameter-distribution}
\vspace{2mm}
\captionof{table}{
The mean values $\mu_{\rm{par}}$ and standard deviations $\sigma_{\rm{par}}$ of the Gaussian parameter distributions used for the \textbf{2G} training data and the validation data for both \textbf{2G} and \textbf{2U}, and the ranges of the uniform parameter distributions used for the \textbf{2U} training data in the two-component case. The resulting parameter distributions are shown in Fig.~\ref{fig:case6-7_parameter-distribution}.}
\label{tab:case6-7_parameter-table}
\scalebox{0.9}{
\begin{tabular}{c c c c}
\hline\hline
\textbf{Gaussian} & Component $i$ & $\mu_{\rm{par}}$ & $\sigma_{\rm{par}}$ \\
\hline
$f_{i}$ [MHz] & 0 (red) & $1.0$ & $0.10$ \\
& 1 (blue) & $1.2$ & $0.10$\\
\hdashline
$\phi_{i}$ [rad] & 0 (red) & $-0.20$ & $0.10$ \\
& 1 (blue) & $0.20$ & $0.10$\\
\hdashline
$\tau_{i}$ [$\mu$s] & 0 (red) & $1.8$ & $0.10$ \\
& 1 (blue) & $2.0$ & $0.10$\\
\hdashline
$A_{i}$ & 0 (red) & $1.0$ & $0.10$ \\
& 1 (blue) & $2.0$ & $0.10$\\
\hline
\textbf{Uniform} & Component $i$ & Minimum & Maximum \\
\hline
$f_{i}$ [MHz] & 0 (red) & $0.80$ & $1.3$ \\
& 1 (blue) & $1.0$ & $1.5$\\
\hdashline
$\phi_{i}$ [rad] & 0 (red) & $-0.60$ & $0.10$\\
& 1 (blue) & $-0.10$ & $0.60$\\
\hdashline
$\tau_{i}$ [$\mu$s] & 0 (red) & $1.4$ & $2.1$ \\
& 1 (blue) & $1.7$ & $2.4$\\
\hdashline
$A_{i}$ & 0 (red) & $0.30$ & $1.6$ \\
& 1 (blue) & $1.4$ & $2.7$\\
\hline\hline
\end{tabular}
}
\end{figure*}

Second, we considered two-component superposed damped sinusoidal signals with training data generated from Gaussian distributions (\textbf{2G}) and uniform distributions (\textbf{2U}). 
Unlike \textbf{2G-sub} and \textbf{2G-opp}, which were designed as particularly challenging two-component settings, \textbf{2G} and \textbf{2U} represent more standard two-component cases and are introduced as complementary benchmarks for comparing Gaussian and uniform training.
The waveforms were generated according to Eq.~(\ref{eq:multi_damped}), and white Gaussian noise with standard deviation $\sigma_{\rm{noise}} = 1/2^3$ ($\rm{SNR} \sim 15.7 \rm{dB}$) was added. 
The training samples were generated from Gaussian and uniform parameter distributions for \textbf{2G} and \textbf{2U}, respectively.
The validation samples were generated from the same Gaussian parameter distributions as the training data for \textbf{2G}.
The range of the uniform distribution in \textbf{2U} was chosen to cover the parameter range of the validation samples.
To avoid making the evaluation overly dependent on the training-distribution settings, the parameter distributions for the training and validation data were slightly shifted relative to each other. 
The parameters used for each distribution are listed in Table~\ref{tab:case6-7_parameter-table}, and the resulting parameter distributions are shown in Fig.~\ref{fig:case6-7_parameter-distribution}. 

\subsubsection{Three-component superposed damped sinusoidal signals: Uniform training}\label{subsubsec:t_three_G_U}
\begin{figure*}[tbp]
  \centering
  \begin{subfigure}{0.45\linewidth}
    \centering
    \includegraphics[width=70mm]{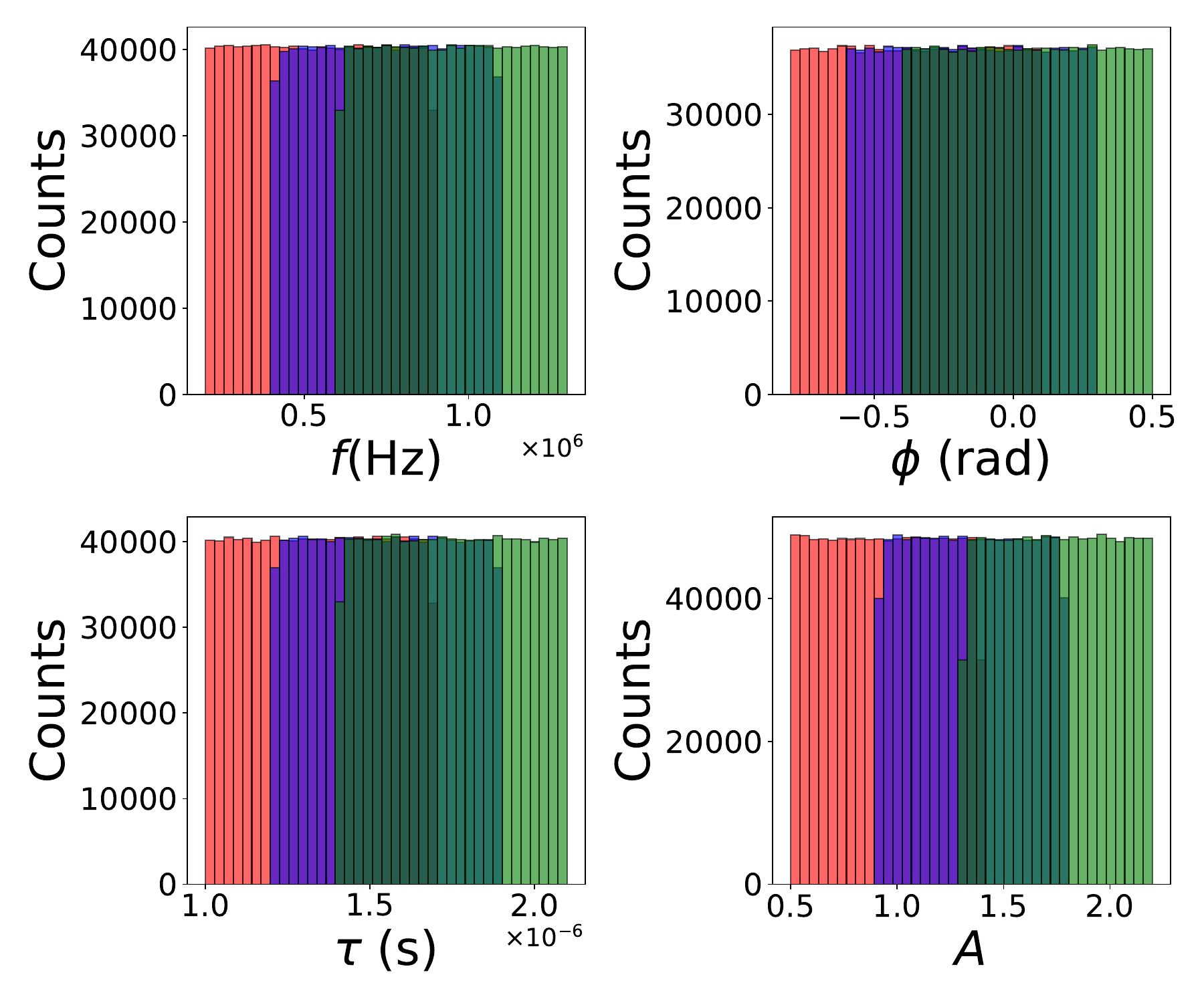}
    \caption{Training data.}
  \end{subfigure}
  \begin{subfigure}{0.45\linewidth}
    \centering
    \includegraphics[width=70mm]{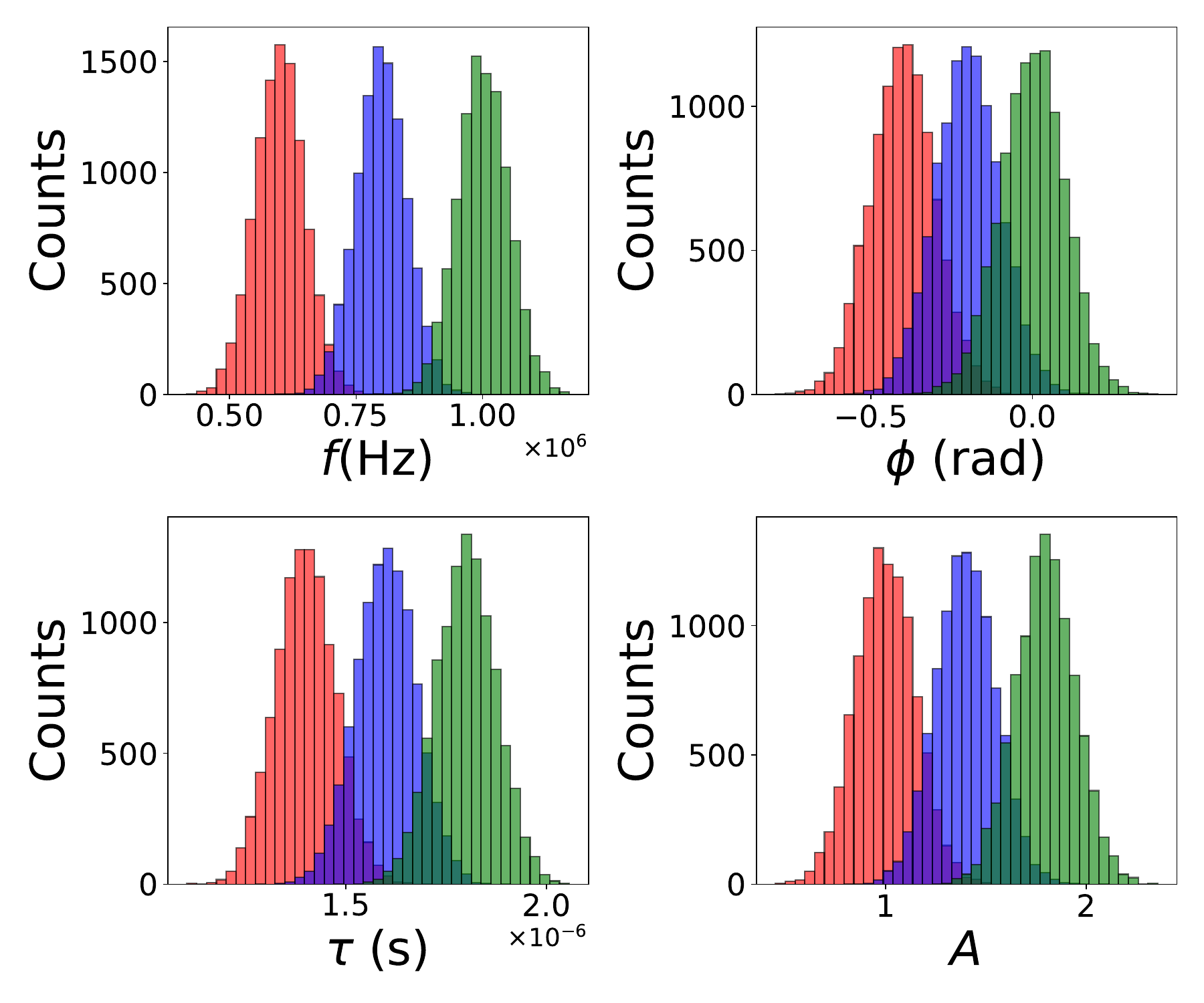}
    \caption{Validation data.}
  \end{subfigure}
  \caption{\textbf{3U}: Training and validation parameter distributions for the three-component case.}
  \label{fig:case8_parameter-distribution}
\vspace{2mm}
\captionof{table}{
Range of uniform parameter distributions used for the training data and the mean values $\mu_{\rm{par}}$ and standard deviations $\sigma_{\rm{par}}$ of the Gaussian parameter distributions used for the validation data for \textbf{3U}. The resulting parameter distributions are shown in Fig.~\ref{fig:case8_parameter-distribution}.}
\label{tab:case8_parameter-table}
\scalebox{0.9}{
\begin{tabular}{c c c c}
\hline\hline
\textbf{Training} & Component $i$ & Minimum & Maximum \\
\hline
$f_{i}$ [MHz] & 0 (red) &$ 0.20$ & $0.90$ \\
& 1 (blue) & $0.40$ & $1.1$\\
& 2 (green) & $0.60$ & $1.3$ \\
\hdashline
$\phi_{i}$ [rad] & 0 (red) & $-0.80$ & $0.10$\\
& 1 (blue) & $-0.60$ & $0.30$\\
& 2 (green) & $-0.40$ & $0.50$ \\
\hdashline
$\tau_{i}$ [$\mu$s] & 0 (red) & $1.0$ & $1.7$ \\
& 1 (blue) & $1.2$ & $1.9$\\
& 2 (green) & $1.4$ & $2.1$ \\
\hdashline
$A_{i}$ & 0 (red) & 0.50 & $1.4$ \\
& 1 (blue) & 0.90 & $1.8$\\
& 2 (green) & $1.3$ & $2.2$ \\
\hline
\textbf{Validation} & Component $i$ & $\mu_{\rm{par}}$ & $\sigma_{\rm{par}}$ \\
\hline
$f_{i}$ [MHz] & 0 (red) & $0.60$ & $0.10$ \\
& 1 (blue) & $0.80$ & $0.10$\\
& 2 (green) & $1.0$ & $0.10$ \\
\hdashline
$\phi_{i}$ [rad] & 0 (red) & $-0.40$ & $0.10$ \\
& 1 (blue) & $-0.20$ & $0.10$\\
& 2 (green) & $0.00$ & $0.10$ \\
\hdashline
$\tau_{i}$ [$\mu$s] & 0 (red) & $1.4$ & $0.10$ \\
& 1 (blue) & $1.6$ & $0.10$\\
& 2 (green) & $1.8$ & $0.10$ \\
\hdashline
$A_{i}$ & 0 (red) & $1.0$ & $0.10$ \\
& 1 (blue) & $1.4$ & $0.10$\\
& 2 (green) & $1.8$ & $0.10$ \\
\hline\hline
\end{tabular}
}
\end{figure*}

Finally, we considered three-component superposed damped sinusoidal signals with training data generated from uniform parameter distributions as \textbf{3U},
in order to examine a more severe setting in which the overlap among the component-wise training distributions is increased.
We focus on the noise level $\sigma_{\rm{noise}} = 1/2^2$ ($\rm{SNR} \sim 11.4 \rm{dB}$), since the multiple superposition makes the parameter estimation difficult.
The parameters used for each distribution are listed in Table~\ref{tab:case8_parameter-table}, and the resulting parameter distributions are shown in Fig.~\ref{fig:case8_parameter-distribution}(a) for the training data and Fig.~\ref{fig:case8_parameter-distribution}(b) for the validation data, respectively. 

\subsection{Evaluation methods}\label{subsec:eval}
The estimation performance was assessed using three complementary metrics. 
These metrics were designed to evaluate the method from three different viewpoints: 
the agreement between the reconstructed and original waveforms, 
the consistency between the true and estimated parameter distributions, 
and the parameter-wise estimation errors.

First, for the validation data, we evaluated the agreement between the original noise-free waveform and the denoised waveform reconstructed by the decoder using the match score~\cite{ref:Pycbc}. 
This quantity provides a measure of waveform similarity while allowing for an overall time shift and phase offset:
\begin{equation}
\mathrm{Match\ score} = \max_{t_{\mathrm{s}}, \phi_{\mathrm{s}}} 
\frac{
    \left( y_{\mathrm{deno}}, y_{\mathrm{org}} \right)
}{
    \left( y_{\mathrm{org}}, y_{\mathrm{org}} \right)^{1/2}
    \left( y_{\mathrm{deno}}, y_{\mathrm{deno}} \right)^{1/2}
}, \label{eq:match}
\end{equation}
It takes values between 0 and 1, with values closer to 1 indicating better agreement between the two waveforms.
Here, the inner product $\left( y_{\mathrm{deno}}, y_{\mathrm{org}} \right)$ between denoised and original waveforms is defined as
\begin{equation}
\left( y_{\mathrm{deno}}, y_{\mathrm{org}} \right) = 4 \int_{0}^{\infty} 
\frac{
    \tilde{Y}_{\mathrm{deno}}(f)\cdot \tilde{Y}^{*}_{\mathrm{org}}(f)
}{
    S_n(f)
} \, df, 
\end{equation}
where $*$ denotes complex conjugation, $\tilde{Y}_{\mathrm{deno}}(f)$ is the frequency-domain representation of the denoised waveform, $\tilde{Y}_{\mathrm{org}}(f)$ is that of the original waveform, and $S_n(f)$ denotes the noise power spectral density. 
Since white Gaussian noise was used in this study, we set $S_n(f)=1$. 
In Eq.~(\ref{eq:match}), $t_s$ and $\phi_s$ denote the time shift and phase offset, respectively, and were optimized to maximize the waveform agreement. 

Second, we compared the distributions of the true parameters in the validation data with those of the estimated parameters obtained from the latent-space representation.
This comparison provides a direct visualization of how accurately the latent-space representation reproduces the underlying parameter distributions.

Third, we quantified the estimation error for each parameter by directly comparing the true and estimated values for each validation sample. 
For a given parameter, this yields one error value for each validation sample. 
The estimation performance was then assessed statistically from the distribution of these sample-wise errors by computing their mean values and standard deviations, which are summarized in the next section.
For the frequency, decay time, and amplitude, we used the relative error,
\begin{equation}
\frac{|p_{\mathrm{est}}-p_{\mathrm{true}}|}{|p_{\mathrm{true}}|},
\end{equation}
while for the phase we used the absolute error,
\begin{equation}
|\phi_{\mathrm{est}}-\phi_{\mathrm{true}}|.
\end{equation}
The phase was treated differently because it is an angular variable, for which a relative error is not a natural measure.

\section{Results and Discussion}\label{sec:results}
In this section, we present and discuss the obtained results. 
The training settings used for each case are summarized in Table~\ref{tab:training_settings}. 
Because the network architecture and parameter-space coverage differ among experiments, the optimization settings were selected separately in preliminary training runs. We chose learning rates that yielded a stable, non-divergent decrease of the training loss and selected the numbers of epochs after the training loss had approximately plateaued. The independently generated validation data were reserved for the final performance evaluation.
The training curves and related diagnostics showed no clear sign of overfitting. 
Below, we present the validation results using the evaluation metrics introduced in Section~\ref{subsec:eval}.

\begin{table}[tb]
\centering
\caption{Training settings for \textbf{2G-sub}--\textbf{3U}.}
\label{tab:training_settings}
\begin{tabular}{c c c c}
\hline\hline
Case & Learning rate & Encoder epochs & Decoder epochs \\
\hline
\textbf{2G-sub} & 0.001 & 350 & 200 \\
\textbf{2G-opp} & 0.001 & 500 & 500 \\
\textbf{5G} & 0.005 & 350 & 200 \\
\textbf{1G} & 0.001 & 350 & 400 \\
\textbf{1U} & 0.001 & 350 & 400 \\
\textbf{2G} & 0.0001& 250 & 150 \\
\textbf{2U} & 0.0001& 250 & 150 \\
\textbf{3U} & 0.0001& 250 & 150 \\
\hline\hline
\end{tabular}
\end{table}

\subsection{Two-component superposed damped sinusoidal signals: Gaussian training}\label{subsubsec:r_two_G}
We first present the results for the two-component cases with Gaussian training, focusing on the challenging settings introduced in Section~\ref{subsec:data_generation}.

\begin{figure*}[tbp]
  \centering
  \begin{subfigure}{0.6\linewidth}
  \includegraphics[width=95mm]{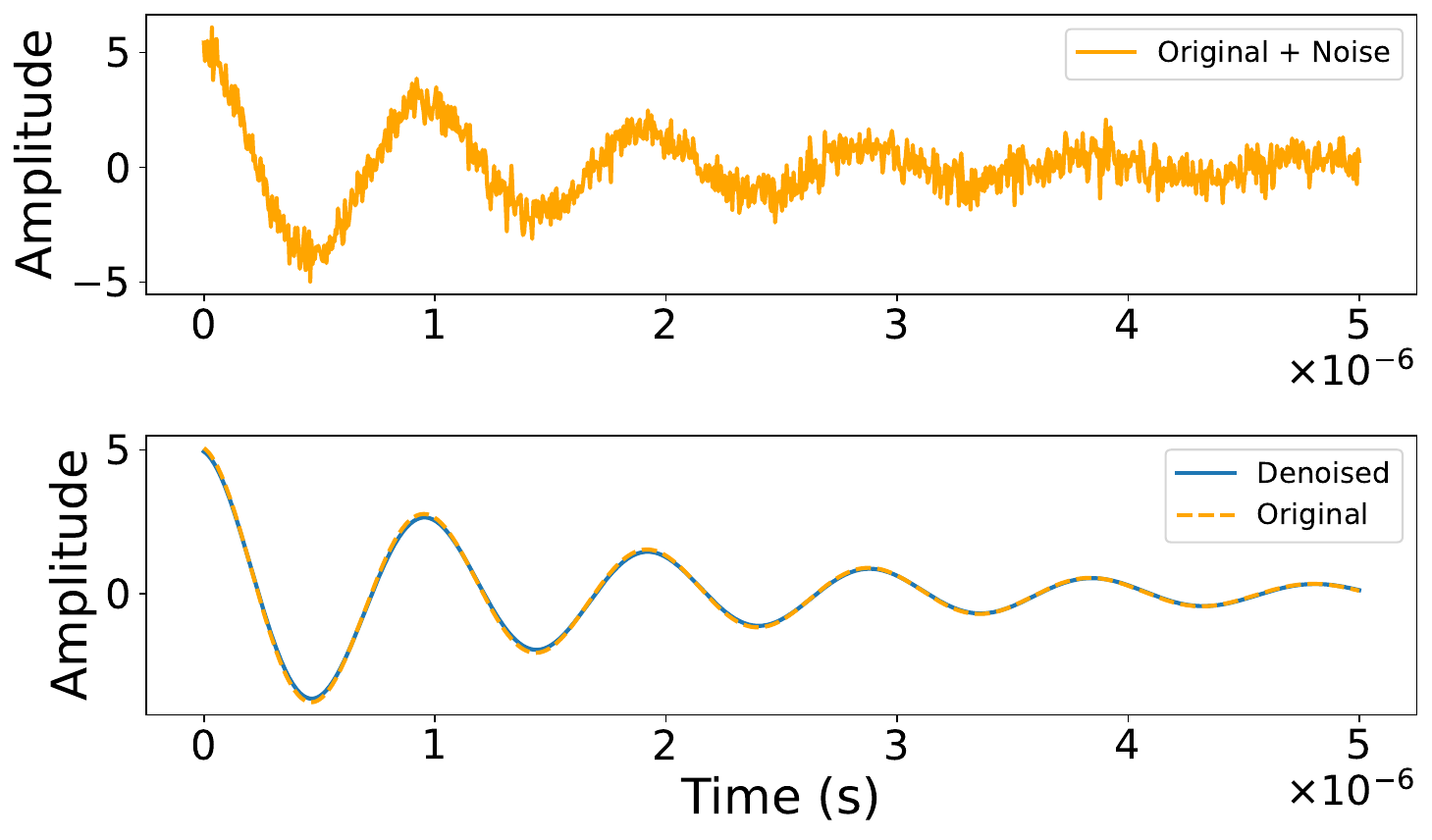}
   \caption{Waveforms before and after denoising (Match score: 1.000).}
   \end{subfigure}
  \begin{subfigure}{0.6\linewidth}
  \includegraphics[width=95mm]{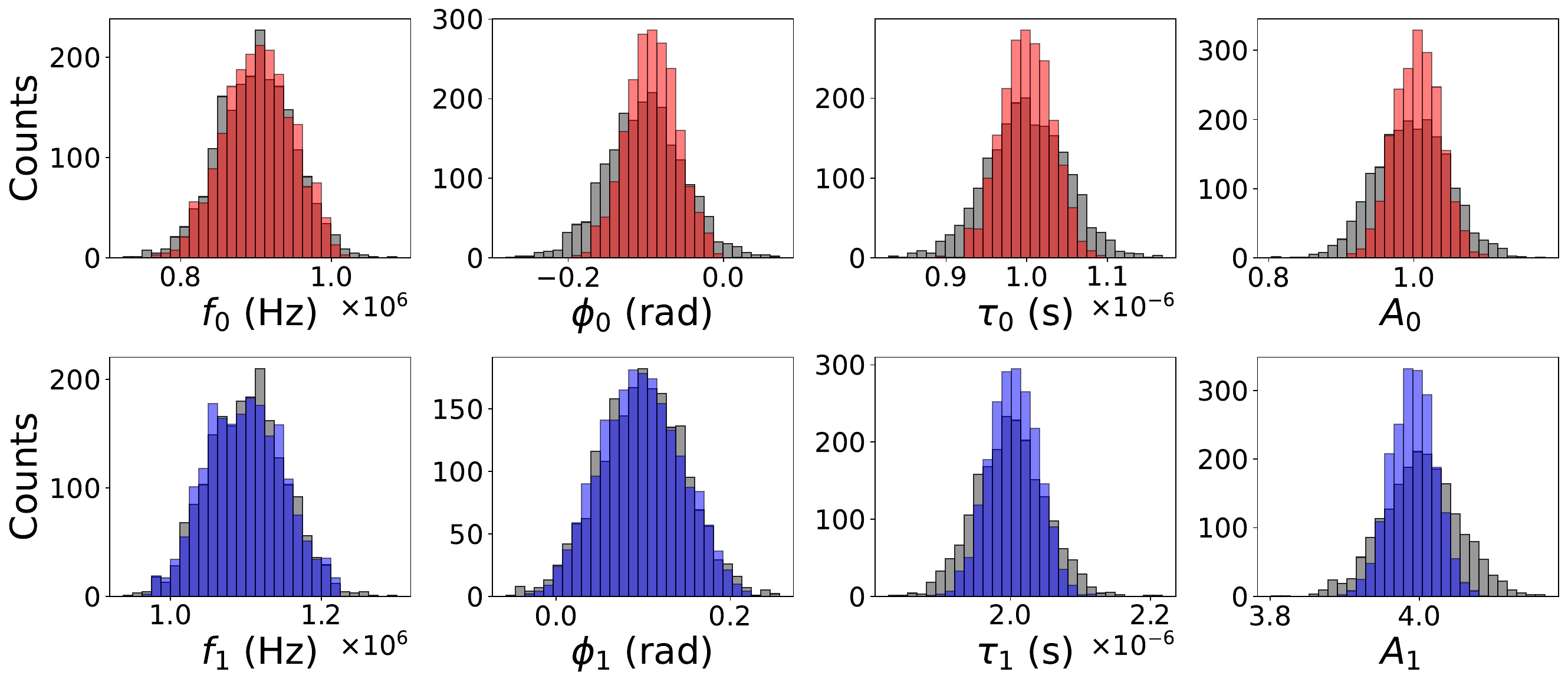}
   \caption{Distributions of true and estimated parameters. Gray histograms indicate the true parameter distributions, and red and blue histograms indicate the estimated parameter distributions for components 0 and 1, respectively.}
   \end{subfigure}
  \caption{Results for \textbf{2G-sub}.}
  \label{fig:case1_result}
\end{figure*}

\begin{figure*}[tbp]
  \centering
   \begin{subfigure}{0.6\linewidth}
   \includegraphics[width=95mm]{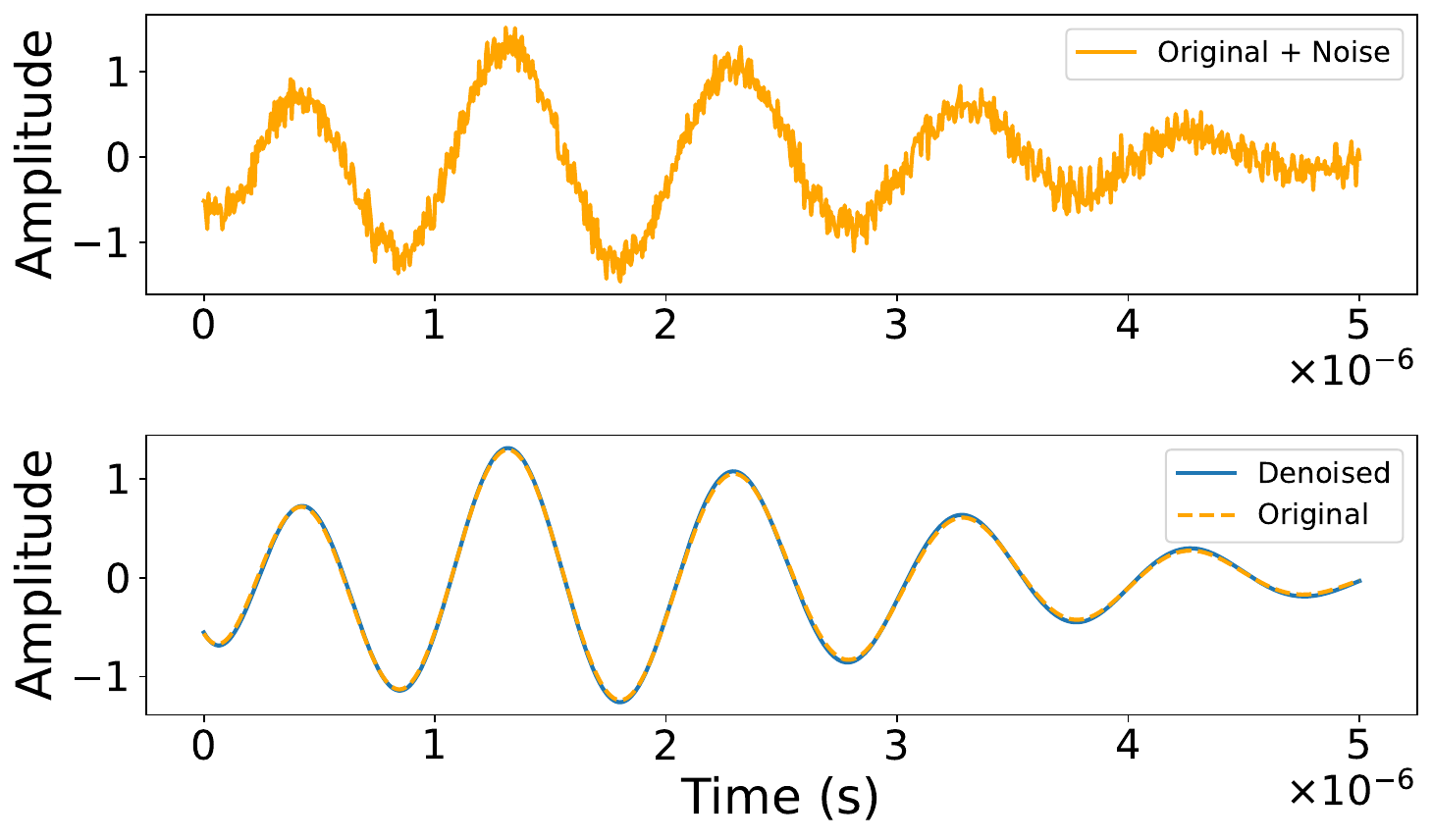}
    \caption{Waveforms before and after denoising (Match score: 0.999).}
   \end{subfigure}
    \begin{subfigure}{0.6\linewidth}
  \includegraphics[width=95mm]{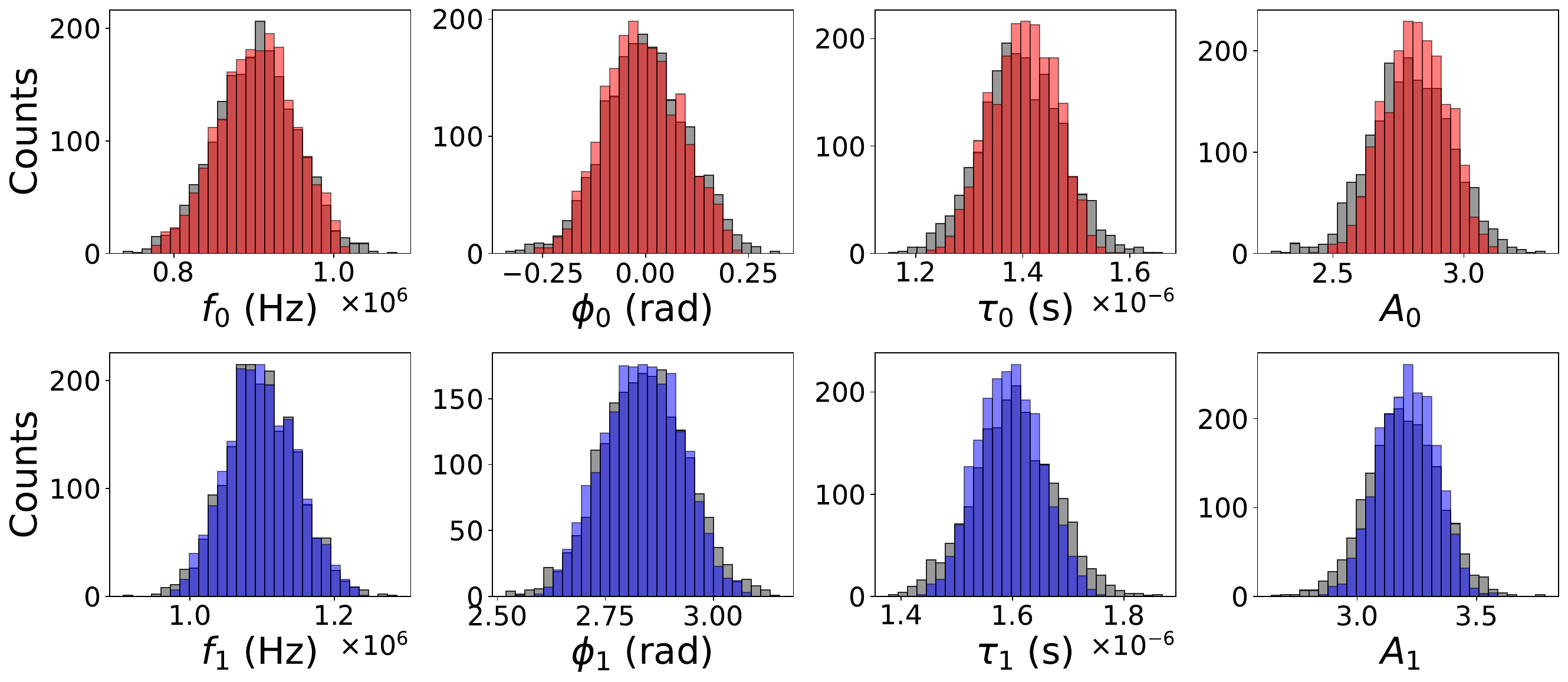}
    \caption{Distributions of the true and estimated parameters. The colors follow the same convention as in Fig.~\ref{fig:case1_result}.}
   \end{subfigure}
 \caption{Results for \textbf{2G-opp}.}
  \label{fig:case2_result}
\vspace{2mm}
\captionof{table}{
Mean values and standard deviations of the match score, the relative errors (rel.) in frequency, decay time, and amplitude, and the absolute error (abs.) in phase for the validation data in \textbf{2G-sub} and \textbf{2G-opp}.
} \label{tab:case1_table}
\scalebox{0.90}{
\begin{tabular}{c | c | c c c c c }
\hline\hline
 & Match score & $i$ & $f_i$ (rel.) & $\phi_i$ [rad] (abs.) & $\tau_i$ (rel.) & $A_i$ (rel.) \\ 
\hline
\textbf{2G-sub}
& $0.999 \pm 0.006$ 
& $0$
& $0.025 \pm 0.020$
& $0.046 \pm 0.034$
& $0.044 \pm 0.033$
& $0.046 \pm 0.035$ \\
& & $1$
& $0.006 \pm 0.005$
& $0.021 \pm 0.016$
& $0.019 \pm 0.015$
& $0.012 \pm 0.009$ \\ \hline
\textbf{2G-opp}
& $0.998 \pm 0.009$ 
& $0$
& $0.007 \pm 0.007$
& $0.036 \pm 0.031$
& $0.028 \pm 0.022$
& $0.033 \pm 0.026$ \\
& & $1$
& $0.004 \pm 0.005$
& $0.032 \pm 0.027$
& $0.023 \pm 0.018$
& $0.027 \pm 0.022$ \\
\hline\hline
\end{tabular}
}
\vspace{2mm}
\captionof{table}{
 Mean values and standard deviations of the match score, the relative errors (rel.) in frequency, decay time, and amplitude, and the absolute error (abs.) in phase for the validation data in \textbf{2G-sub} and \textbf{2G-opp} for the simple MP method~\cite{ref:MP,ref:MP2}.
} \label{tab:case1_table_MP}
\scalebox{0.90}{
\begin{tabular}{c | c | c c c c c }
\hline\hline
 & Match score & $i$ & $f_i$ (rel.) & $\phi_i$ [rad] (abs.) & $\tau_i$ (rel.) & $A_i$ (rel.) \\ 
\hline
\textbf{2G-sub}
& $0.860 \pm 0.288$ 
& $0$
& $3.3\times10^{1} \pm 3.8 \times 10^{1}$
& $1.4 \pm 1.1$
& $ 1.3\times10^{1}  \pm 1.7\times 10^{2}$
& $0.74 \pm 0.33$ \\
& & $1$
& $7.1 \pm 2.4 \times 10^{1}$
& $0.31 \pm 0.78$
& $0.43 \pm 6.4$
& $0.27 \pm 0.27$ \\ \hline
\textbf{2G-opp}
& $0.998 \pm 0.027$ 
& $0$
& $0.15 \pm 2.6$
& $0.15 \pm 0.49$
& $0.058 \pm 0.12$
& $0.50 \pm 0.086$ \\
& & $1$
& $0.22 \pm 4.3$
& $0.12 \pm 0.39$
& $0.074 \pm 1.0$
& $0.50 \pm 0.078$ \\
\hline\hline
\end{tabular}
}
\end{figure*}

For \textbf{2G-sub}, which contains a rapidly decaying, low-amplitude component, Fig.~\ref{fig:case1_result}(a) shows an example of the input signal and the corresponding denoised waveform. 
The lower panel of Fig.~\ref{fig:case1_result}(a) shows that the denoised waveform agrees well with the original waveform, successfully reproducing both the amplitude variations and oscillation patterns. 
For the validation data, Table~\ref{tab:case1_table} summarizes the mean values and standard deviations of the match score and of the sample-wise parameter-estimation errors.
The mean match score was $0.999$ with a standard deviation of $0.006$, indicating excellent reconstruction performance. 
Because the match score is bounded above by 1, this standard deviation reflects the spread of a distribution accumulated near the upper bound rather than a symmetric uncertainty interval. 
In particular, the expression $0.999 \pm 0.006$ does not imply that any individual sample has a match score exceeding $1$.
A slight degradation in the match score was observed for samples located near the edges of the parameter distributions.

For \textbf{2G-opp}, in which the two components have nearly opposite phases, the results are summarized in Fig.~\ref{fig:case2_result} and Table~\ref{tab:case1_table}. 
As in \textbf{2G-sub}, the denoised waveform reproduces the original waveform well. 
When the frequency difference between the two components was small, the waveform shapes became more similar and the cancellation effect due to the nearly opposite phases became more pronounced, leading to a reduction in the match score for some samples.
Nevertheless, a high overall match score of $0.998 \pm 0.008$ was achieved. 

For both \textbf{2G-sub} and \textbf{2G-opp}, the distributions of the estimated parameters agree well with those of the true parameters, as shown in Fig.~\ref{fig:case1_result}(b) and Fig.~\ref{fig:case2_result}(b). 
Table~\ref{tab:case1_table} summarizes the relative errors for the frequency, decay time, and amplitude, as well as the absolute error for the phase.\footnote{The error values reported here for \textbf{2G-sub} and \textbf{2G-opp} are not identical to those in~\cite{ref:Iida2026}. In the previous proceedings paper, signed relative errors $\frac{p_{\mathrm{est}}-p_{\mathrm{true}}}{p_{\mathrm{true}}}$ were used, whereas in the present paper we use absolute relative errors $\frac{|p_{\mathrm{est}}-p_{\mathrm{true}}|}{|p_{\mathrm{true}}|}$ so that the tables directly quantify the magnitude of the estimation error.}
Overall, the parameter-estimation accuracy is high in both cases. 

For \textbf{2G-opp}, we also examined higher-noise settings, such as $\sigma_{\rm{noise}} = 1/2$ and $1/2^2$. 
In those cases, both the match score and the parameter-estimation accuracy tended to deteriorate. 
This is consistent with the fact that \textbf{2G-opp} is particularly sensitive to noise because the superposed signal can be reduced by partial cancellation between the two components. 

A more detailed investigation under realistic noise conditions will be left for future work.

We emphasize that \textbf{2G-sub} and \textbf{2G-opp} were selected as particularly challenging two-component examples. 
For other two-component settings, we likewise observed high match scores and good agreement between the true and estimated parameters.
These results indicate that the present method achieves performance comparable to, and in some respects better than, that reported previously for single-component damped sinusoidal signals~\cite{ref:autoencoder}.

To provide a reference against a conventional parametric signal-processing technique, we applied the simple MP method~\cite{ref:MP,ref:MP2} to the same validation datasets used for \textbf{2G-sub} and \textbf{2G-opp}. 
For each noisy validation waveform in the two-component experiments, we constructed a $(N_t-L)\times(L+1)$ Hankel matrix, set $L=N_t/2$, and retained $p=4$ singular vectors, corresponding to two conjugate pole pairs. The discrete poles were obtained as the eigenvalues of $V_1^{\dagger}V_2$; positive-frequency poles were retained, and their complex coefficients were estimated by a least-squares fit of a Vandermonde matrix. Frequencies, decay times (using $\tau=-1/\alpha$, where $\alpha$ is the damping rate), amplitudes, and phases were then obtained from the poles and coefficients. Component 0 was matched to the component with the smaller estimated amplitude in the subdominant setting (\textbf{2G-sub}) and the smaller estimated phase in the opposite-phase setting (\textbf{2G-opp}).

Tables~\ref{tab:case1_table} and~\ref{tab:case1_table_MP} summarize the parameter estimation errors obtained by the proposed autoencoder and the MP method. For \textbf{2G-sub}, the proposed method achieved smaller errors for the subdominant component, whose short decay time and weak amplitude make accurate estimation difficult. The MP method frequently exhibited larger deviations in the decay time and amplitude estimates because the contribution of the weaker component was easily masked by noise.
For \textbf{2G-opp}, where the two components have nearly opposite phases and partially cancel each other, the proposed method also outperformed the MP method. In particular, the MP estimates became unstable when destructive interference reduced the effective signal amplitude, whereas the latent-space representation learned by the autoencoder remained relatively robust.
These results suggest that the proposed method provides more stable parameter estimation than the simple MP method under challenging noise conditions, especially when weak or partially cancelled components are present. 

In the remaining subsection, we focus our report on the results of the autoencoder.

%

\subsection{Five-component superposed damped sinusoidal signal: Gaussian training}\label{subsubsec:r_five_G}
\begin{figure*}[tbp]
  \centering
   \begin{subfigure}{0.5\linewidth}
   \includegraphics[width=0.9\linewidth]{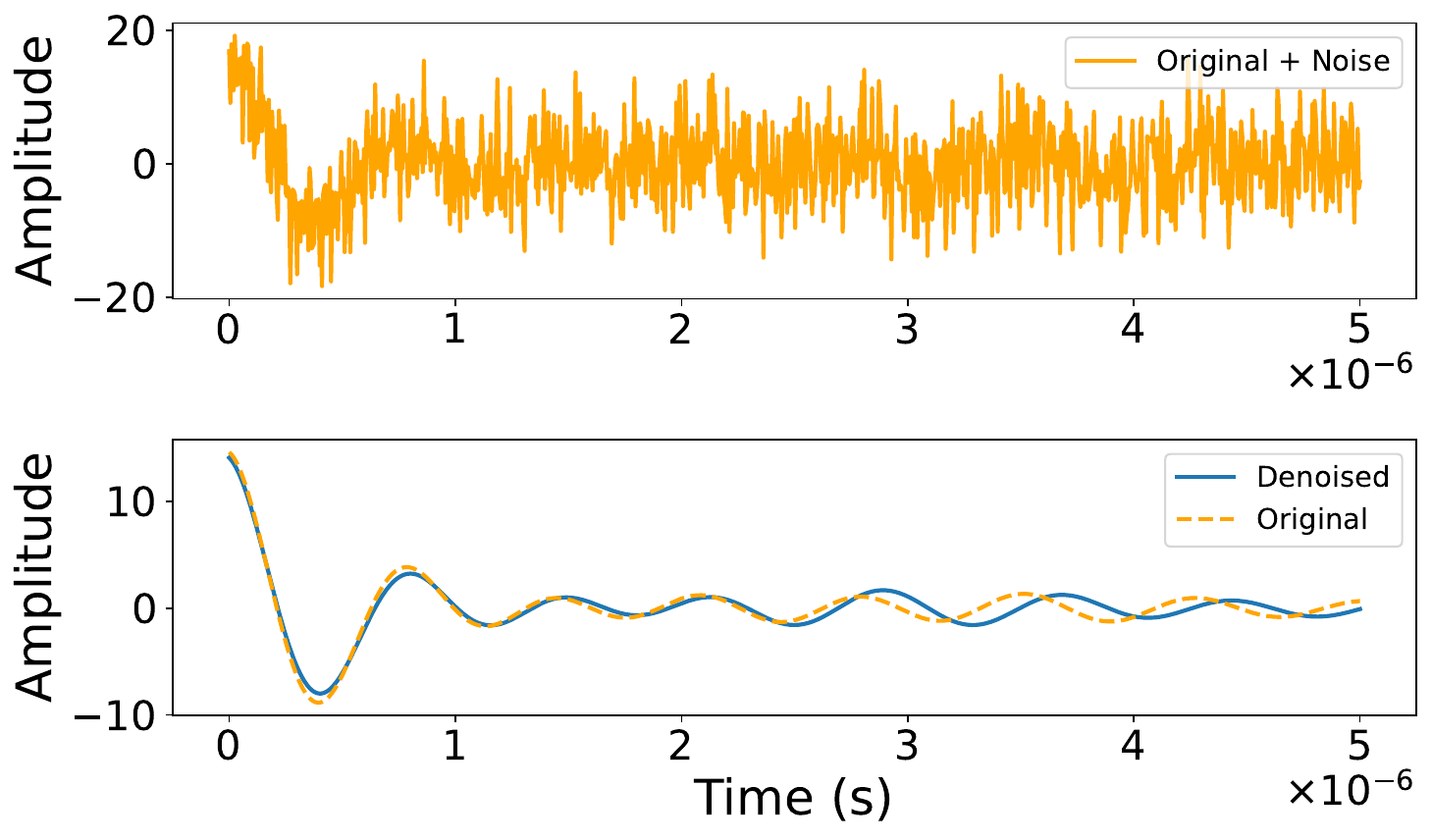}
    \caption{Waveforms before and after denoising (Match score: 0.973).}
   \end{subfigure}
   \begin{subfigure}{0.5\linewidth}
   \includegraphics[width=0.9\linewidth]{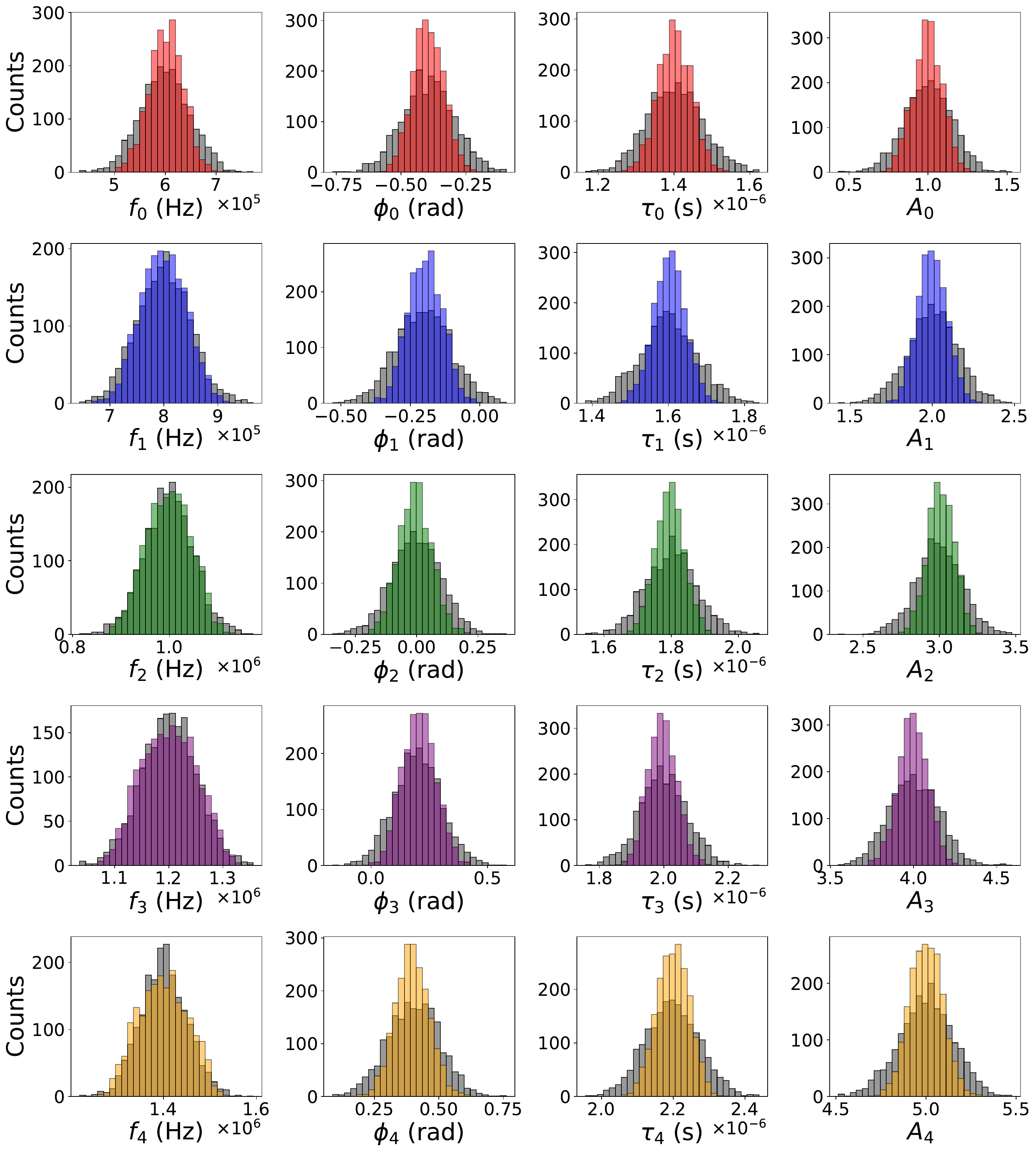}
   \caption{Distributions of the true and estimated parameters. Gray histograms indicate the true parameter distributions, and colored histograms indicate the estimated parameter distributions for components 0--4.}
   \end{subfigure}
 \caption{Results for \textbf{5G}.}
   \label{fig:case3_result}
\end{figure*}

We next present the results for the five-component case with Gaussian training, \textbf{5G}. 
Figure~\ref{fig:case3_result}(a) shows an example of the input signal and the corresponding denoised waveform.
The denoised waveform agrees well with the original waveform, reproducing both the amplitude variation and the oscillatory structure.

\begin{figure*}[tbp]
  \centering
   \begin{subfigure}{0.5\linewidth}
 \includegraphics[width=0.9\linewidth]{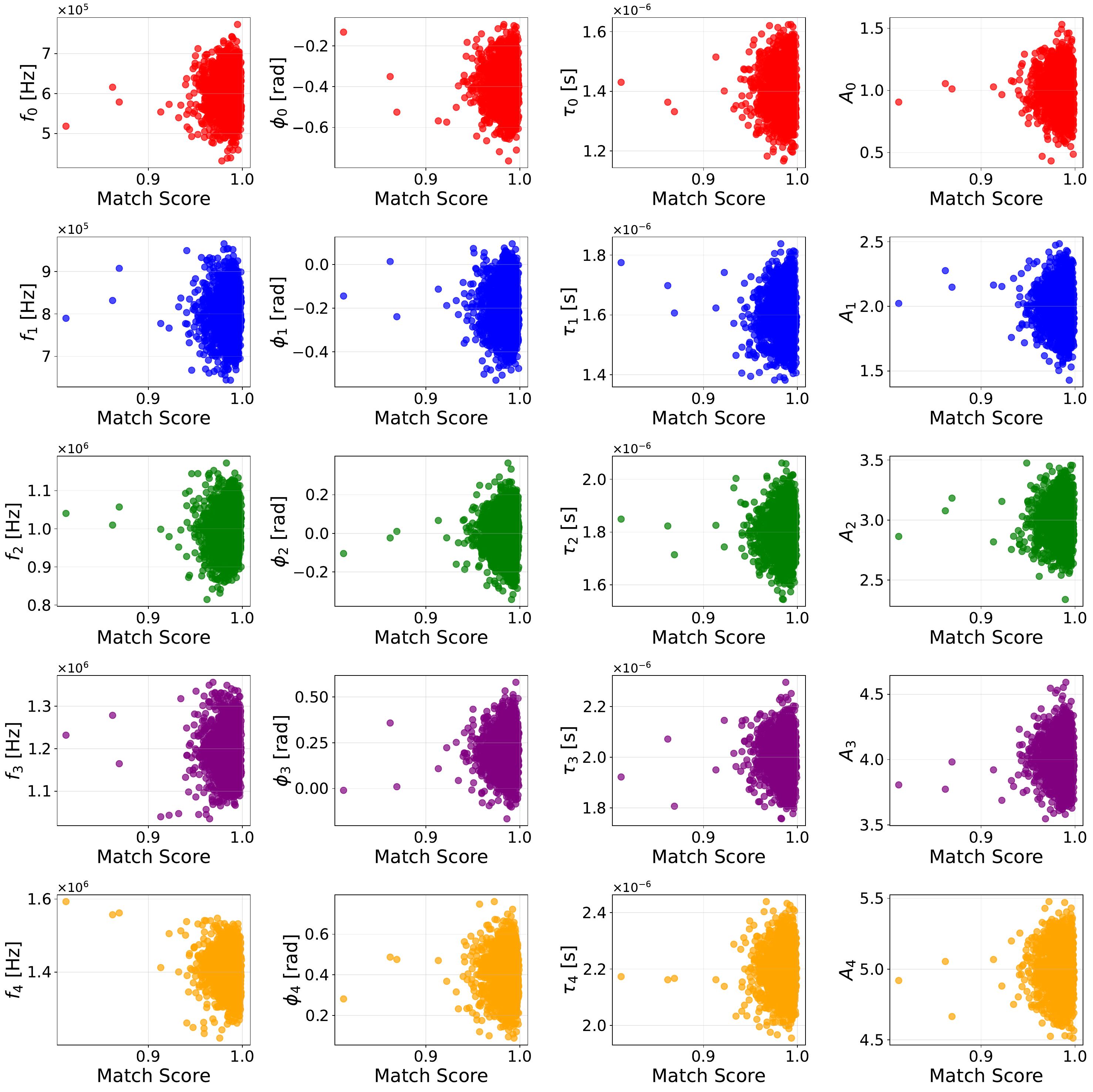}
\end{subfigure}
 \caption{Scatter plots of the match score against each parameter for \textbf{5G}.}
   \label{fig:case3_result_scatter}
\vspace{2mm}
\captionof{table}{Mean values and standard deviations of the match score, the relative errors (rel.) in frequency, decay time, and amplitude, and the absolute error (abs.) in phase for the validation data in \textbf{5G}. }
\label{tab:case3_table}
\scalebox{0.85}{
\begin{tabular}{c | c | c c c c c}
\hline\hline
& Match score & $i$ & $f_i$ (rel.) & $\phi_i$ [rad] (abs.) & $\tau_i$ (rel.) & $A_i$ (rel.) \\
\hline
\textbf{5G} 
& $0.985\pm 0.012$
& $0$
& $0.072 \pm 0.056$
& $0.093 \pm 0.069$
& $0.049 \pm 0.037$
& $0.143 \pm 0.127$ \\
& & $1$
& $0.045 \pm 0.036$
& $0.093 \pm 0.068$
& $0.042 \pm 0.033$
& $0.068 \pm 0.056$ \\
& & $2$
& $0.031 \pm 0.024$
& $0.088 \pm 0.067$
& $0.039 \pm 0.030$
& $0.046 \pm 0.036$ \\
& & $3$
& $0.021 \pm 0.017$
& $0.089 \pm 0.065$
& $0.034 \pm 0.026$
& $0.035 \pm 0.026$ \\
& & $4$
& $0.014 \pm 0.011$
& $0.083 \pm 0.061$
& $0.032 \pm 0.024$
& $0.028 \pm 0.021$ \\
\hline\hline
\end{tabular}
}
\end{figure*}

Figure~\ref{fig:case3_result}(b) shows the distributions of the true and estimated parameters, while Table~\ref{tab:case3_table} summarizes the relative errors for the frequency, decay time, and amplitude, together with the absolute error for the phase. 
Overall, the true and estimated parameter distributions agree well, indicating that accurate parameter estimation was achieved even in the five-component case.

In particular, component 0, which has the smallest amplitude and is buried by the other dominant components over the entire time interval (see Fig.~\ref{fig:case3_parameter-distribution}), is expected to be the most difficult to estimate. 
Nevertheless, the estimation accuracy for this component remains reasonably good.

Figure~\ref{fig:case3_result_scatter} displays a scatter plot of the match score against each parameter. 
For component 4, which is the dominant component, the match score tends to decrease in the high-frequency region. 
This trend is likely due to the fact that the training data were generated from Gaussian parameter distributions, which provide fewer samples in the tails of the distributions, particularly in the high-frequency tail.
Despite this tendency, the overall match score remains high, demonstrating that the proposed method can effectively estimate signals with multiple superposed damped sinusoidal components.

%

\subsection{Gaussian vs Uniform training}\label{sebsec:gauss_uniform}
As we discussed above, for more general situations, in which the true parameter values are not known a priori, it is natural to train the encoder and decoder using uniform parameter distributions for each component of the damped sinusoidal signals. 
We present the corresponding results and discussion below, comparing Gaussian vs uniform training.

\subsubsection{Single-component damped sinusoidal signals: Gaussian vs Uniform training}\label{subsubsec:r_single_G_U}
We next compare the effect of the training-data distribution on the generalization performance in the parameter estimation of a single damped sinusoidal signal. 
Here, we present the results for \textbf{1G} and \textbf{1U}. 

\begin{figure*}[tbp]
  \centering
   \begin{subfigure}{0.6\linewidth}
   \includegraphics[width=95mm]{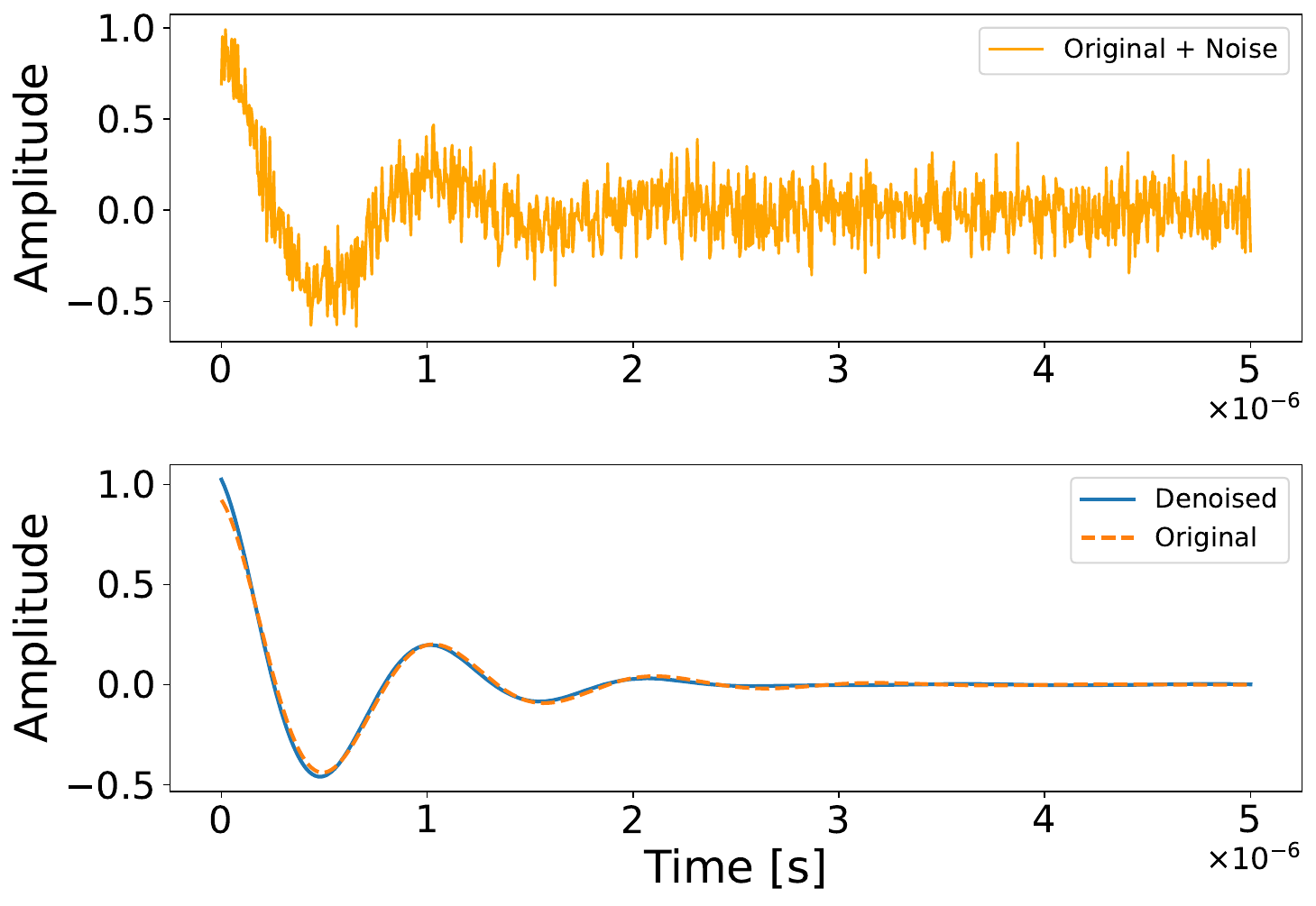}
    \caption{Waveforms before and after denoising (Match score: 0.998).}
   \end{subfigure}
    \begin{subfigure}{0.6\linewidth}
  \includegraphics[width=95mm]{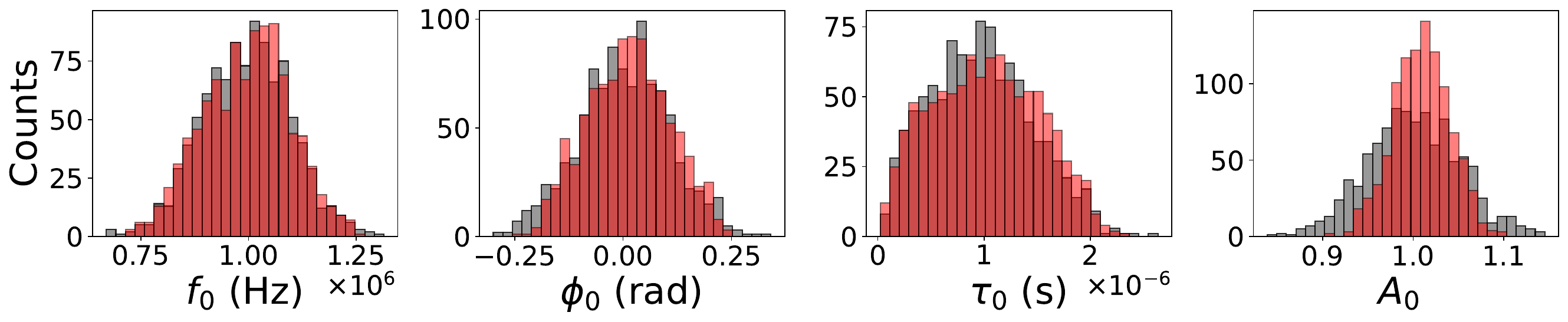}
    \caption{Distributions of true and estimated parameters. Gray histograms indicate the true parameter distributions, and red histograms indicate the estimated parameter distributions.}
   \end{subfigure}
 \caption{Results for \textbf{1G}.}
  \label{fig:case4_result}
\end{figure*}

\begin{figure*}[tbp]
  \centering
   \begin{subfigure}{0.6\linewidth}
   \includegraphics[width=95mm]{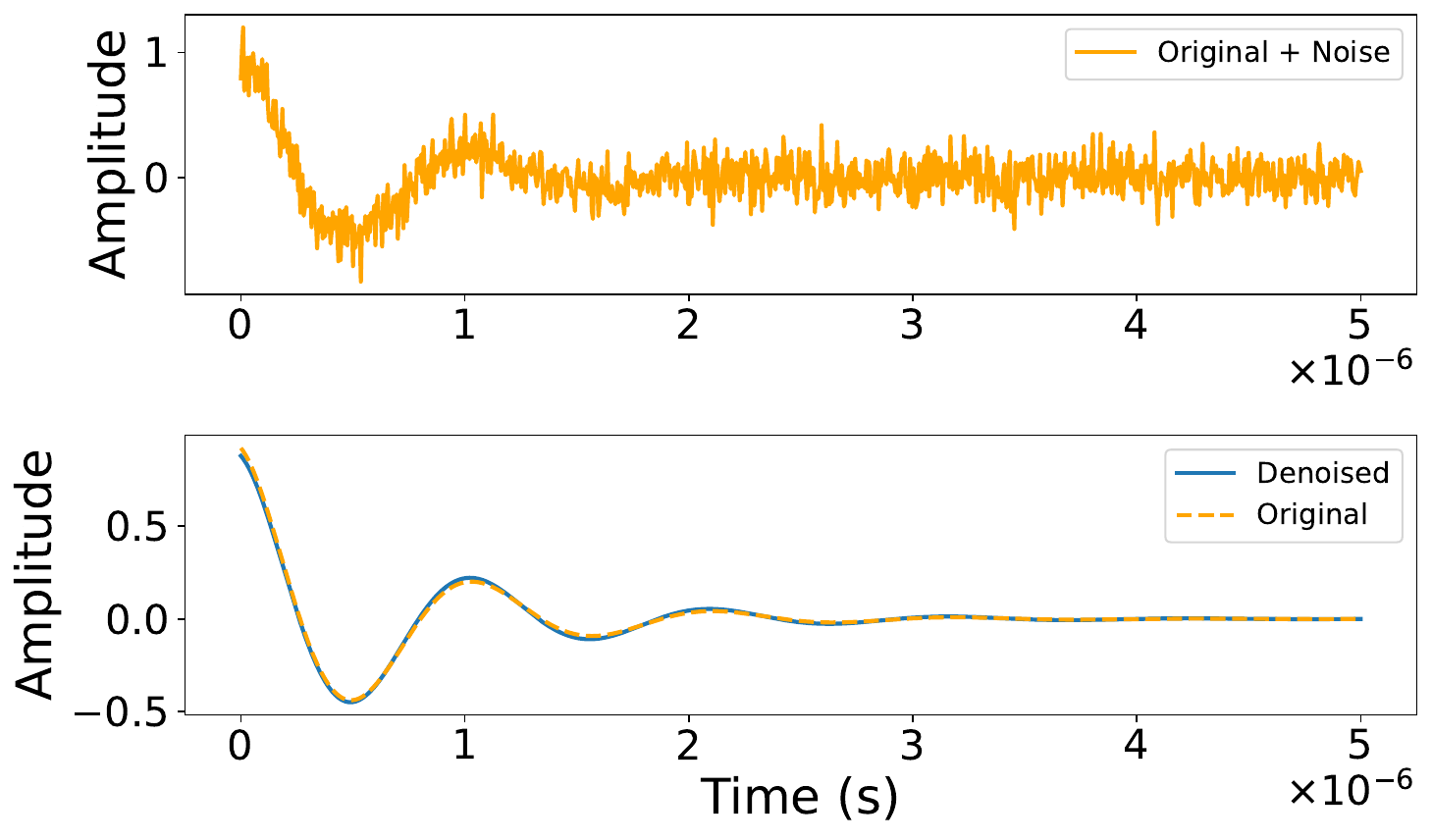}
    \caption{Waveforms before and after denoising (Match score: 0.997).}
   \end{subfigure}
    \begin{subfigure}{0.6\linewidth}
  \includegraphics[width=95mm]{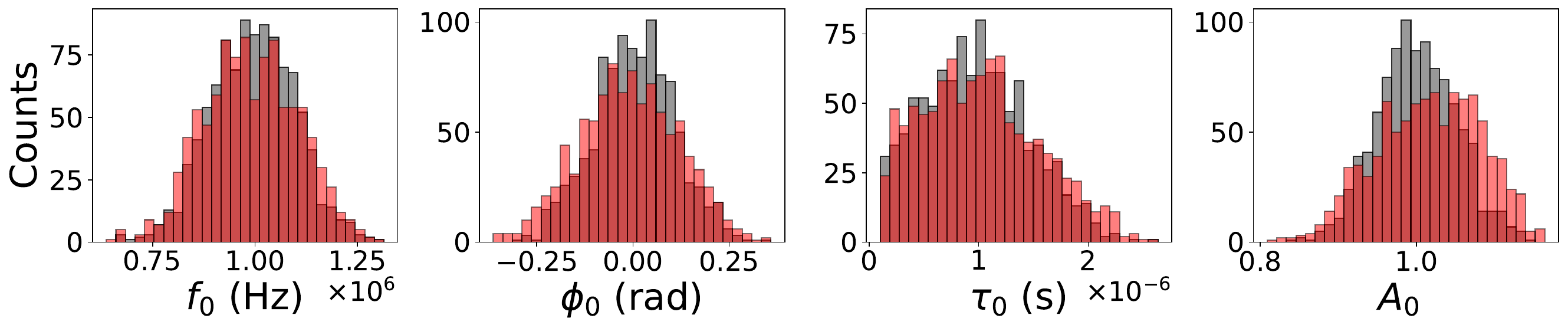}
    \caption{Distributions of the true and estimated parameters. The colors follow the same convention as in Fig.~\ref{fig:case4_result}.}
   \end{subfigure}
 \caption{Results for \textbf{1U}.}
  \label{fig:case5_result}
\end{figure*}

\begin{figure*}[tbp]
  \centering
  \includegraphics[width=0.8\linewidth]{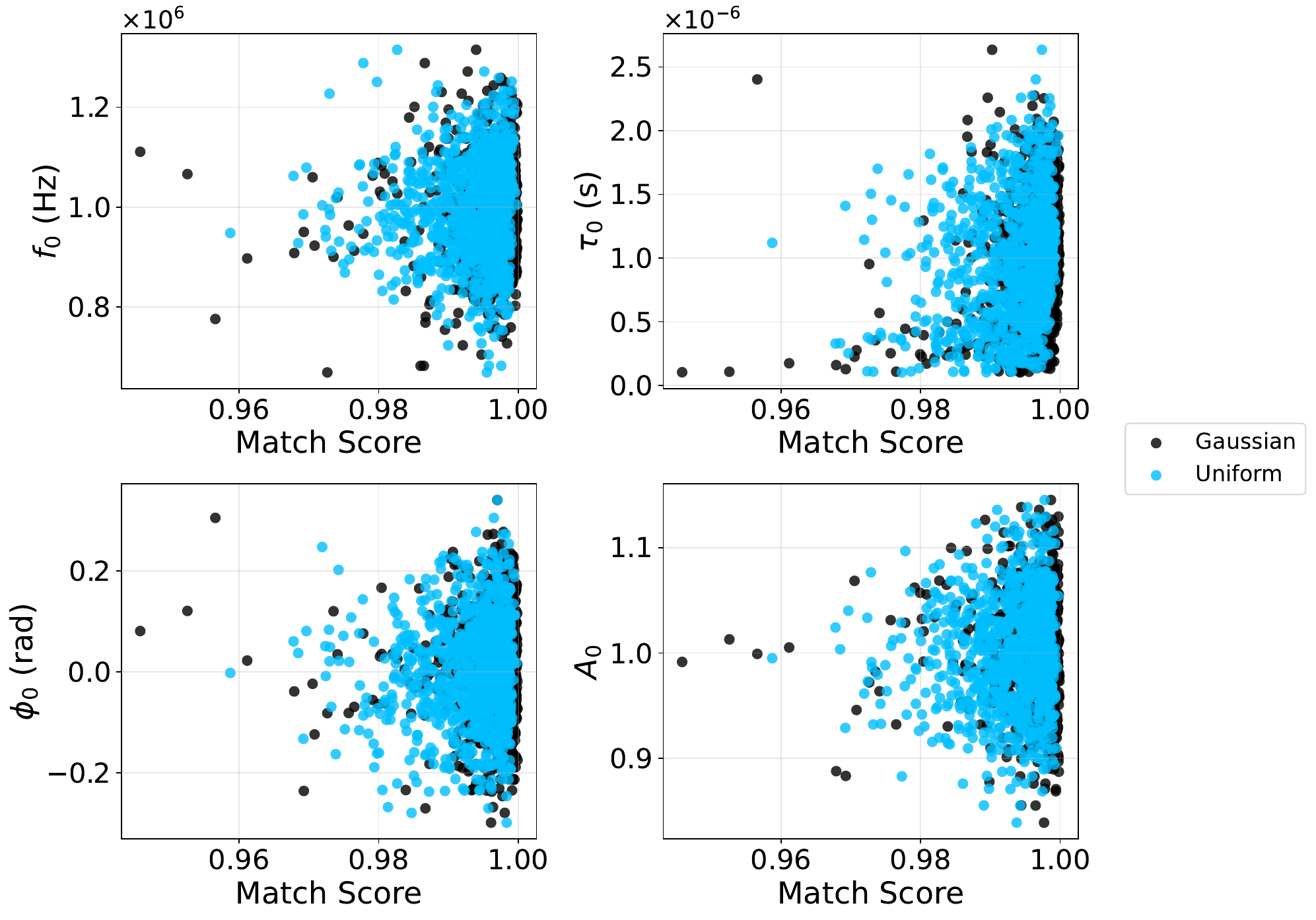}
  \caption{Comparison of match scores for \textbf{1G} (black) and \textbf{1U} (cyan).}
  \label{fig:case4-5_match}
\vspace{2mm}
\captionof{table}{
Mean values and standard deviations of the match score, the relative errors (rel.) in frequency, decay time, and amplitude, and the absolute error (abs.) in phase for the validation data in \textbf{1G} and \textbf{1U}. 
} \label{tab:case4-5_table}  
\scalebox{0.90}{
\begin{tabular}{c | c |c c c c}
\hline\hline
 & Match score &$f_0$ (rel.) & $\phi_0$ [rad] (abs.)  & $\tau_0$ (rel.)  & $A_0$ (rel.) \\ 
\hline
\textbf{1G}
& $0.996 \pm 0.005$
& $0.021 \pm 0.028$
& $0.040 \pm 0.036$
& $0.103 \pm 0.119$
& $0.046 \pm 0.035$ \\ \hline
\textbf{1U}
& $0.993 \pm 0.006$
& $0.030 \pm 0.040$
& $0.057 \pm 0.050$
& $0.099 \pm 0.095$
& $0.061 \pm 0.044$ \\
\hline\hline
\end{tabular}
}
\end{figure*}

Table~\ref{tab:case4-5_table} summarizes the relative errors for the frequency, decay time, and amplitude, together with the absolute error for the phase. 
Figures~\ref{fig:case4_result}(a) and \ref{fig:case5_result}(a) show an example of the input signal and the corresponding denoised waveform for \textbf{1G} and \textbf{1U}, respectively. 
In both cases, the denoised waveform agrees well with the original waveform, indicating that denoising and waveform reconstruction remain effective. 

Figures~\ref{fig:case4_result}(b) and \ref{fig:case5_result}(b) show the distributions of the true and estimated parameters. 
For both \textbf{1G} and \textbf{1U}, no large discrepancy is observed between the true and estimated parameter distributions. 

Figure~\ref{fig:case4-5_match} shows the scatter plots of the match score against each parameter. 
Compared with \textbf{1G}, \textbf{1U}, in which the model was trained with uniformly distributed data, exhibits more uniform performance over the parameter space. 
This result indicates that training with uniform parameter distributions can improve the generalization performance for the single-component case.

\subsubsection{Two-component superposed damped sinusoidal signals: Gaussian vs Uniform training}\label{subsubsec:Two-component}
\begin{figure*}[tbp]
  \centering
  \begin{subfigure}{0.6\linewidth}
    \centering
    \includegraphics[width=\linewidth]{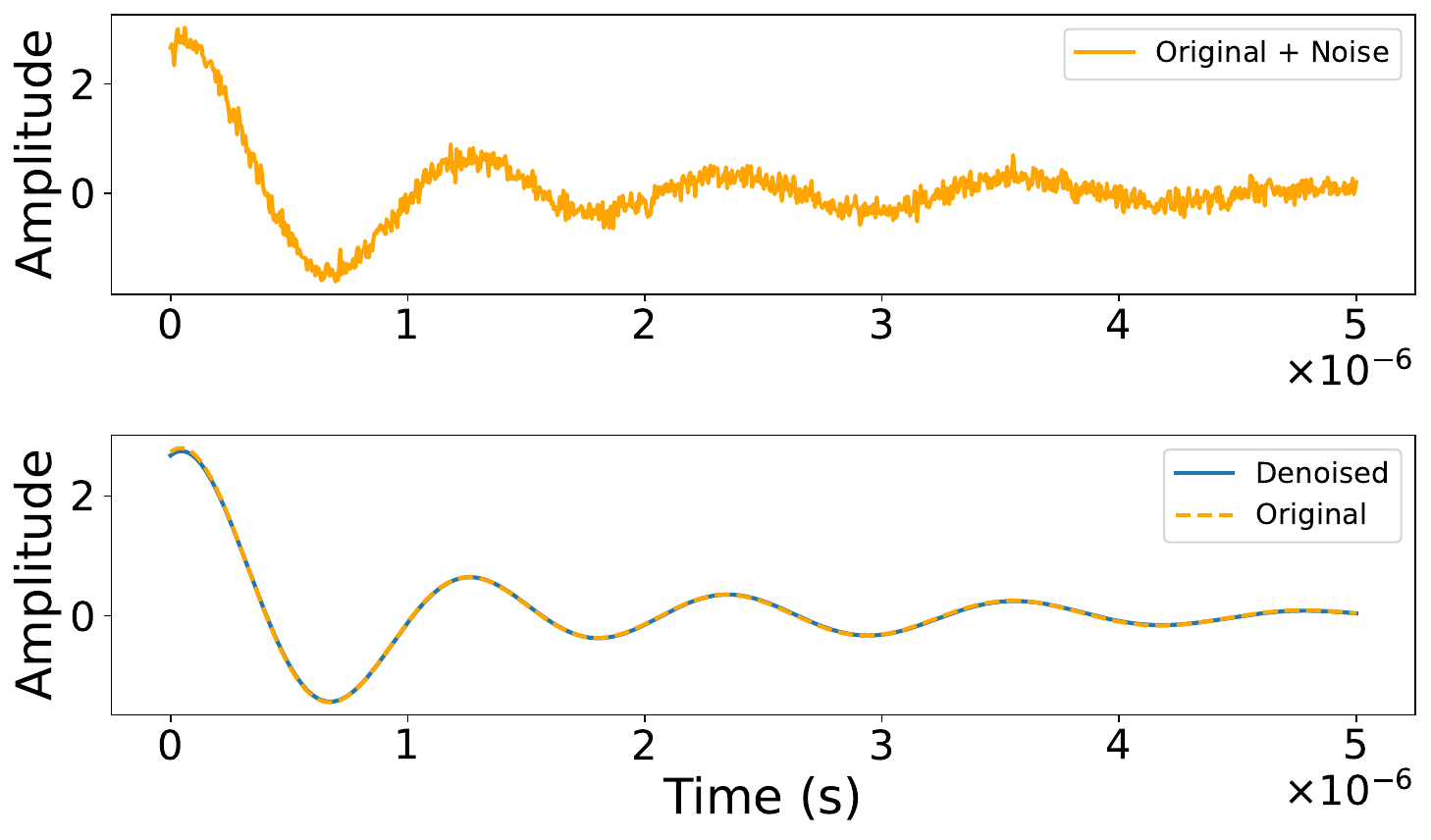}
    \caption{Waveforms before and after denoising (Match score: 1.000).}
  \end{subfigure}
  \begin{subfigure}{0.6\linewidth}
    \includegraphics[width=\linewidth]{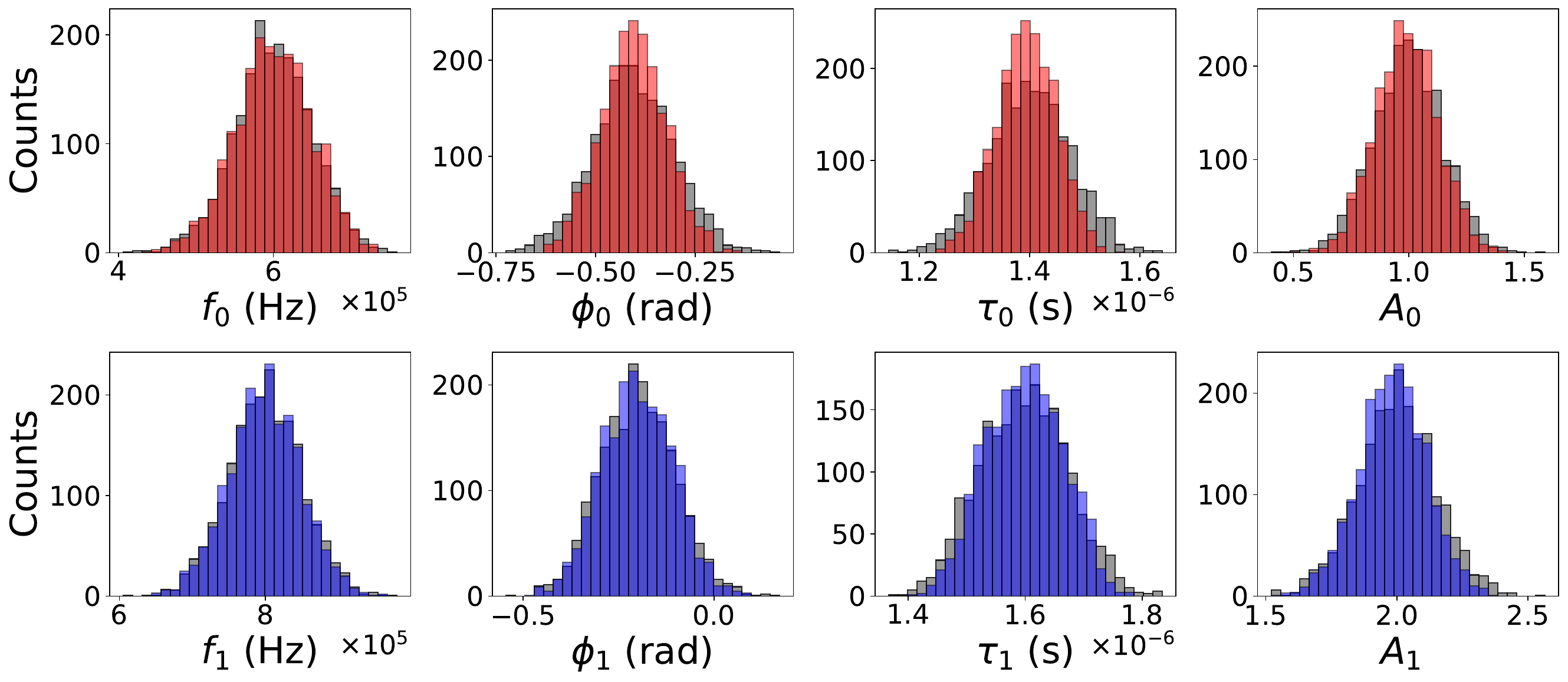}
    \caption{Distributions of the true and estimated parameters. Gray indicates the true parameters, while red and blue indicate the estimated parameters for components 0 and 1, respectively.}
  \end{subfigure}
  \caption{Results for \textbf{2G}.}
  \label{fig:case6_result}
\end{figure*}

\begin{figure*}[tbp]
  \centering
  \begin{subfigure}{0.6\linewidth}
    \centering
    \includegraphics[width=\linewidth]{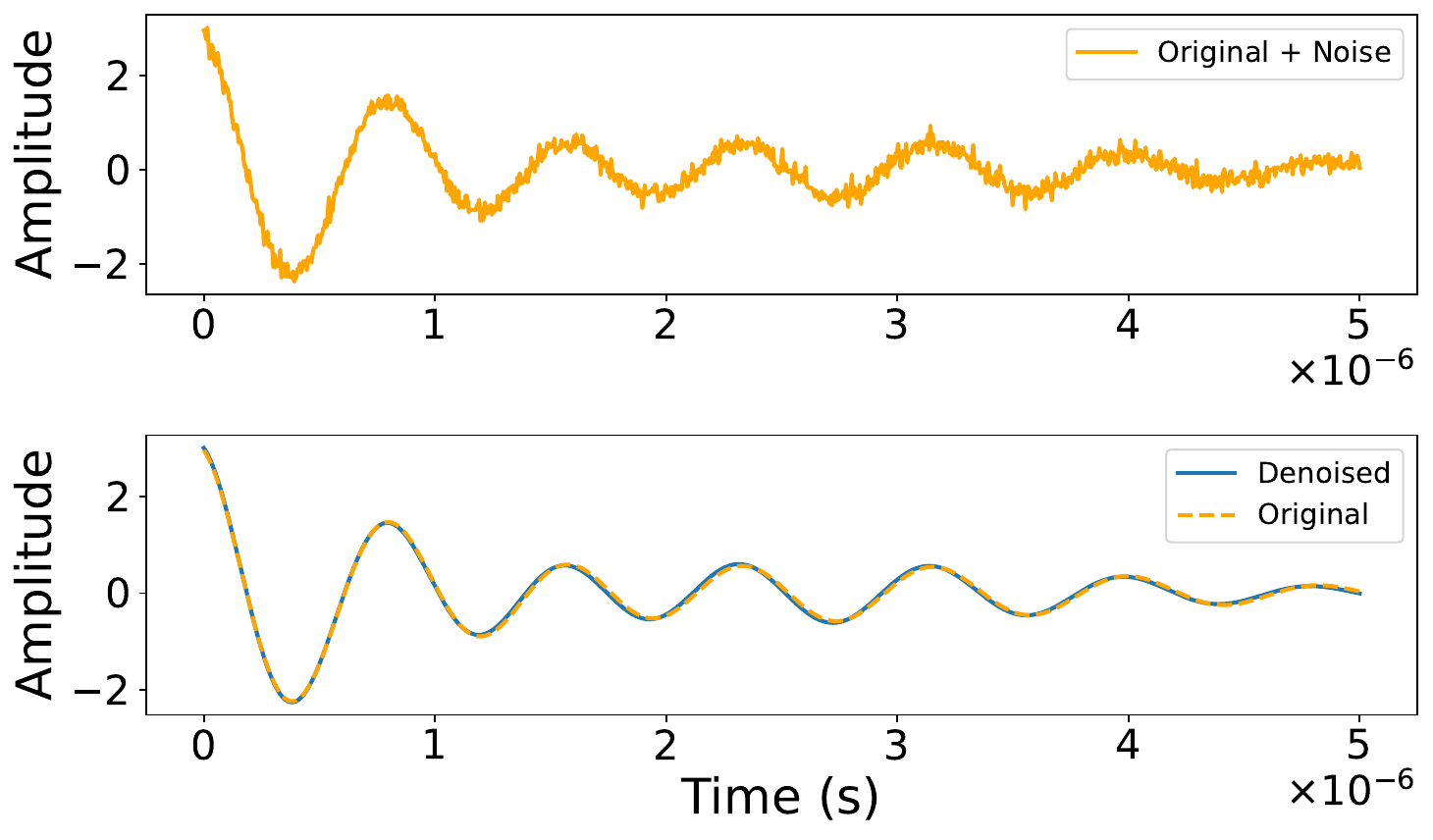}
    \caption{Waveforms before and after denoising (Match score: 0.999).}
  \end{subfigure}
  \begin{subfigure}{0.6\linewidth}
    \includegraphics[width=\linewidth]{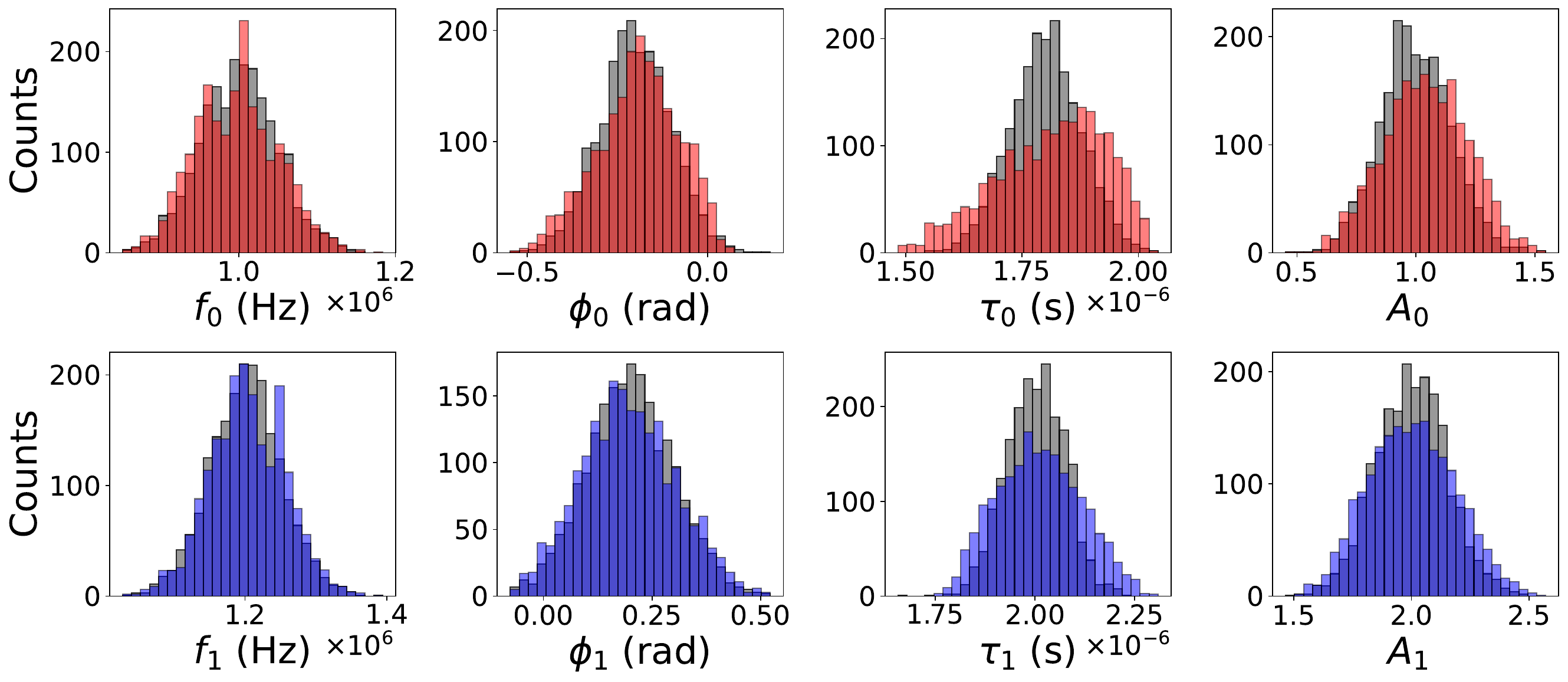}
    \caption{Distributions of the true and estimated parameters. Gray indicates the true parameters, while red and blue indicate the estimated parameters for components 0 and 1, respectively.}
  \end{subfigure}
  \caption{Results for \textbf{2U}.}
  \label{fig:case7_result}
\vspace{2mm}
\captionof{table}{
Mean values and standard deviations of the match score, the relative errors (rel.) in frequency, decay time, and amplitude, and the absolute error (abs.) in phase for the validation data in \textbf{2G} and \textbf{2U}.}
\label{tab:case6-7_table}  
\scalebox{0.90}{
\begin{tabular}{c | c | c c c c c}
\hline\hline
 & Match score & $i$ & $f_i$ (rel.) & $\phi_i$ [rad] (abs.) & $\tau_i$ (rel.) & $A_i$ (rel.) \\
\hline
\textbf{2G}
& $1.000 \pm 0.001$
& $0$
& $0.008 \pm 0.007$
& $0.045 \pm 0.037$
& $0.030 \pm 0.023$
& $0.048 \pm 0.044$ \\
& &
$1$
& $0.003 \pm 0.003$
& $0.023 \pm 0.019$
& $0.018 \pm 0.014$
& $0.023 \pm 0.020$ \\
\hline
\textbf{2U}
& $0.999 \pm 0.001$
& $0$
& $0.008 \pm 0.006$
& $0.053 \pm 0.045$
& $0.043 \pm 0.031$
& $0.078 \pm 0.072$ \\
& &
$1$
& $0.005 \pm 0.004$
& $0.030 \pm 0.026$
& $0.027 \pm 0.021$
& $0.033 \pm 0.025$ \\
\hline\hline
\end{tabular}
}
\end{figure*}

We next present the results for \textbf{2G} and \textbf{2U}. 
Table~\ref{tab:case6-7_table} summarizes the relative errors for the frequency, decay time, and amplitude, together with the absolute error for the phase. 
Figures~\ref{fig:case6_result}(a) and \ref{fig:case7_result}(a) show an example of the input signal and the corresponding denoised waveform for \textbf{2G} and \textbf{2U}, respectively. 
In both cases, the denoised waveform agrees well with the original waveform, indicating that denoising and waveform reconstruction remain effective. 
Figures~\ref{fig:case6_result}(b) and \ref{fig:case7_result}(b) compare the distributions of the true and estimated parameters. 
Although some discrepancies are visible in \textbf{2U}, the overall agreement remains reasonable, indicating that parameter estimation still performs satisfactorily even when the training distribution is less informative. 
At the same time, in \textbf{2U}, a somewhat larger deviation is observed for $\tau_i$ than for the other parameters. 
This trend suggests that overlap among the component-wise training distributions, particularly those of $\tau_i$, makes the separation of the two components more difficult and thereby leads to a larger deviation in the estimated $\tau_i$ distributions. 
This interpretation motivates the three-component uniform-training case discussed next.

\subsubsection{Three-component superposed damped sinusoidal signals: Uniform training}\label{subsubsec:Three-component}
Finally, we present the results for \textbf{3U}. 
Motivated by the two-component comparison above, we consider a three-component model with uniform training as a more severe setting, in which the overlap among the component-wise training distributions is increased. 
By contrast, the Gaussian-training case already performs well even for the more complex five-component setting of \textbf{5G}, so an additional three-component Gaussian benchmark is not expected to change the qualitative conclusion.

\begin{figure*}[tbp]
  \centering
  \begin{subfigure}{0.6\linewidth}
    \centering
    \includegraphics[width=\linewidth]{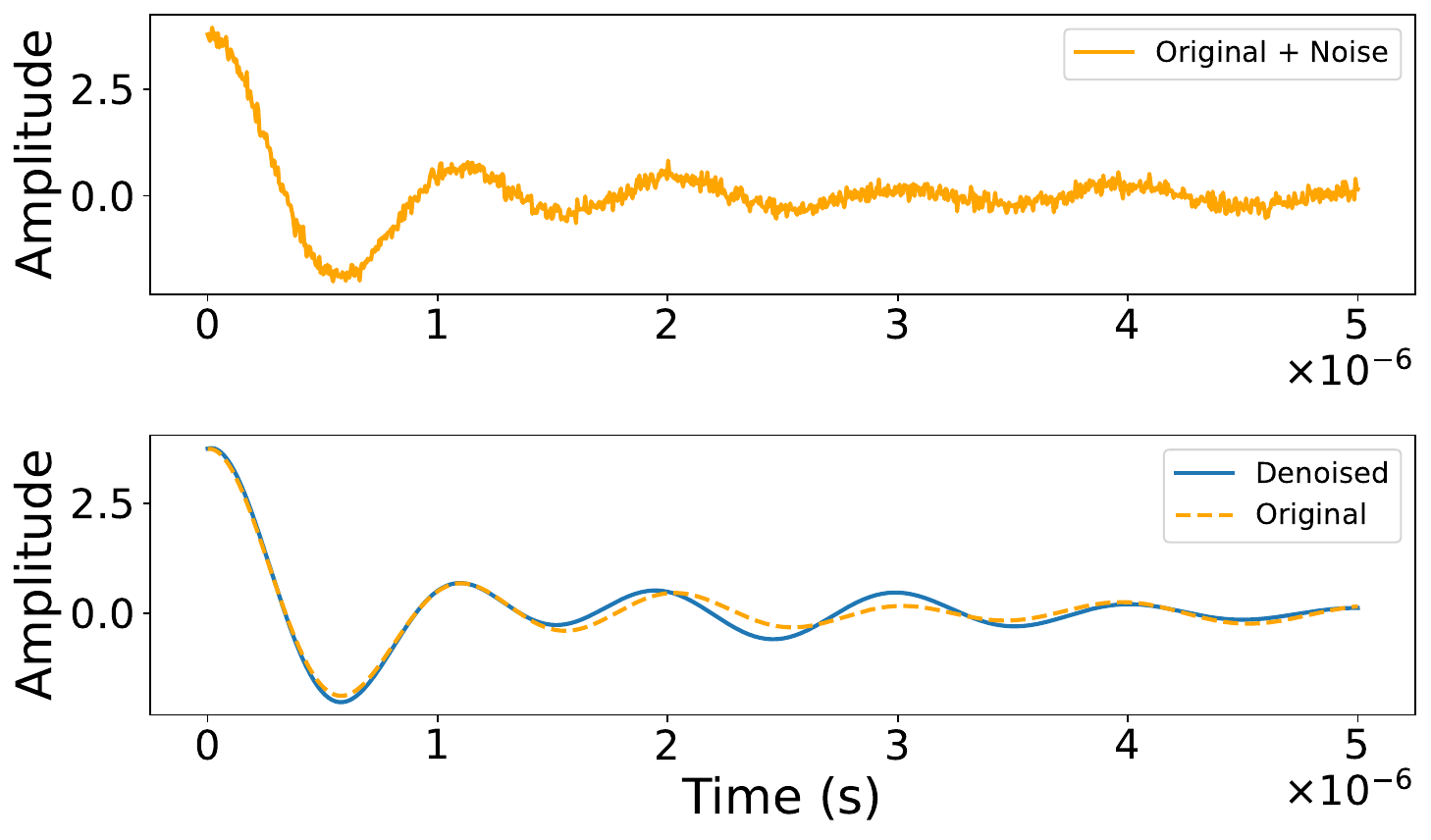}
    \caption{Waveforms before and after denoising (Match score: 0.987).}
  \end{subfigure}
  \begin{subfigure}{0.6\linewidth}
    \includegraphics[width=\linewidth]{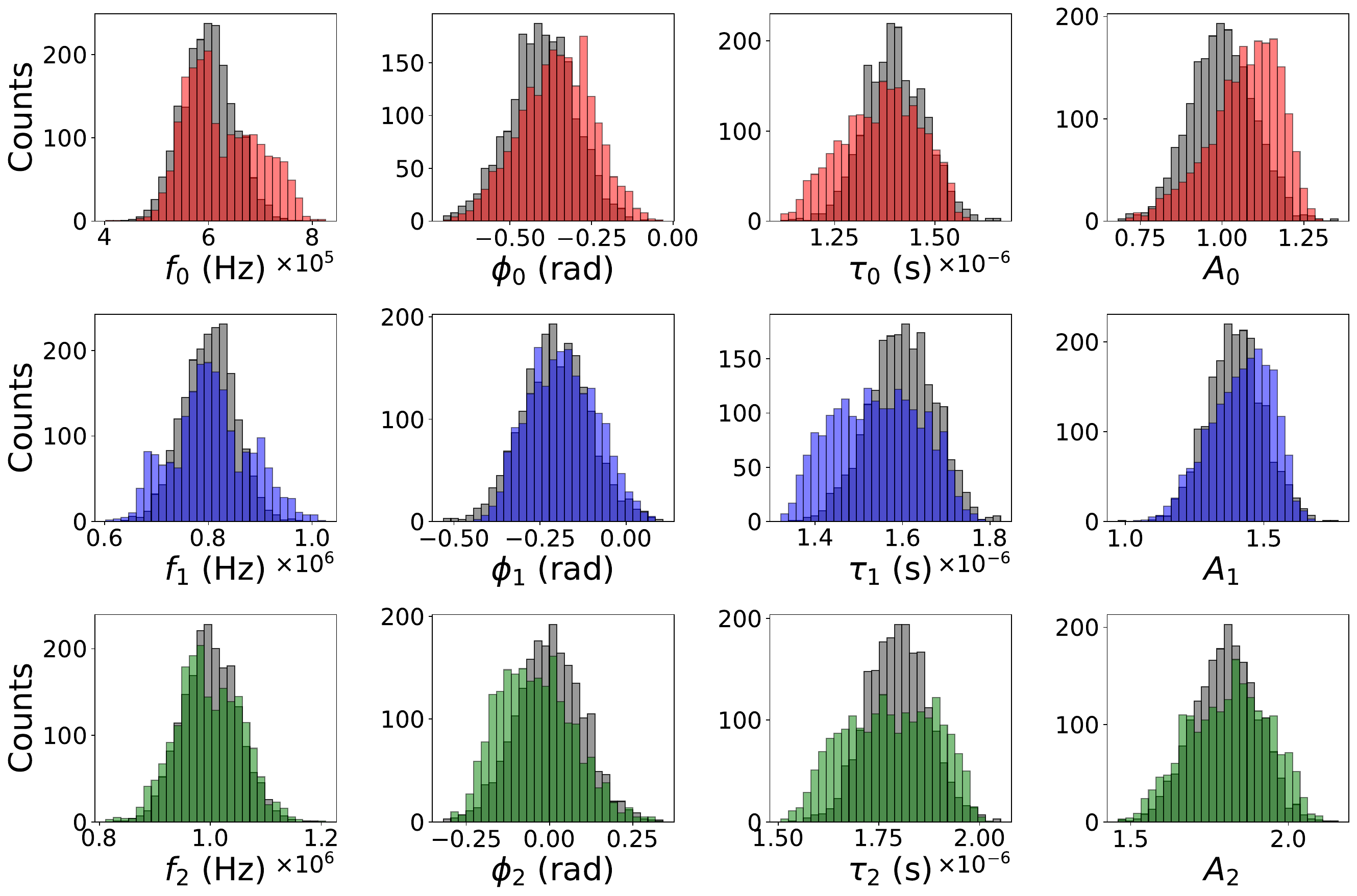}
    \caption{Distributions of the true and estimated parameters. Gray indicates the true parameters, while red, blue and green indicate the estimated parameters for components 0, 1 and 2, respectively.}
  \end{subfigure}
  \caption{Results for \textbf{3U}.}
  \label{fig:case8_result}
\vspace{2mm}
\captionof{table}{Mean values and standard deviations of the match score, the relative errors (rel.) in frequency, decay time, and amplitude, and the absolute error (abs.) in phase for the validation data in \textbf{3U}.}
\label{tab:case8_table}
\scalebox{0.90}{
\begin{tabular}{c | c | c c c c c}
\hline\hline
 & Match score & $i$ & $f_i$ (rel.) & $\phi_i$ [rad] (abs.) & $\tau_i$ (rel.) & $A_i$ (rel.) \\
\hline
\textbf{3U}
& $0.989\pm 0.013$
& $0$
& $0.055 \pm 0.046$
& $0.086 \pm 0.064$
& $0.056 \pm 0.042$
& $0.098 \pm 0.070$ \\
& &
$1$
& $0.041 \pm 0.030$
& $0.081 \pm 0.062$
& $0.056 \pm 0.040$
& $0.055 \pm 0.042$ \\
& &
$2$
& $0.019 \pm 0.015$
& $0.062 \pm 0.046$
& $0.044 \pm 0.032$
& $0.040 \pm 0.030$ \\
\hline\hline
\end{tabular}
}
\end{figure*}

Table~\ref{tab:case8_table} summarizes the relative errors for the frequency, decay time, and amplitude, together with the absolute error for the phase.
Figure~\ref{fig:case8_result}(a) shows an example of the input signal and the corresponding denoised waveform. 
The denoised waveform agrees with the original waveform, indicating that denoising and waveform reconstruction remain effective in this setting. 
Figure~\ref{fig:case8_result}(b) compares the distributions of the true and estimated parameters. 
There is a tendency for the distribution shift to become slightly larger compared to the case with two components. 
This trend supports the interpretation that increased overlap among the component-wise training distributions makes the separation and extraction of individual components more difficult.

\section{Summary}\label{sec:summary}
Building on the previous studies~\cite{ref:autoencoder,ref:Iida2026}, we developed an autoencoder-based method for estimating the frequency, phase, decay time, and amplitude of each component in superposed multi-component damped sinusoidal signals with white Gaussian noise. 
The present work was designed as a baseline study under controlled synthetic settings, with the goal of assessing the basic performance of the method and identifying several observed limitations.
We investigated both two-component and five-component cases and showed that the parameters can be estimated with high accuracy, as summarized in Tables~\ref{tab:case1_table} and \ref{tab:case3_table}. 
The results indicate that the proposed method achieves performance comparable to, and in some cases better than, that previously reported for single-component damped sinusoidal signals~\cite{ref:autoencoder}.
Comparison with the simple MP method demonstrated that the proposed autoencoder yielded more stable parameter estimation for noisy two-component signals, particularly in the presence of subdominant components and partial phase cancellation.

It is particularly noteworthy that the method remains effective even in difficult two-component settings, such as a subdominant component with rapid decay and small amplitude, or two components with nearly opposite phases. 
These are precisely the types of situations in which conventional parameter estimation can become difficult, while at the same time they may arise naturally in physical systems. 
The present results therefore suggest that autoencoder-based parameter estimation may provide a useful tool for investigating the physics behind such signals, including possible future applications to resonant phenomena in damped oscillation systems~\cite{ref:Motohashi_2025}.

The present results also reveal several limitations of the method. 
The estimation accuracy is not uniform over the entire parameter space, and some degradation is observed near the edges of the parameter distributions used for training. 
In the five-component case, a reduction in the match score was found in the high-frequency tail, suggesting that the performance is affected by the limited number of training samples in sparsely populated regions of parameter space. 
Moreover, cases with strong partial cancellation, such as nearly opposite-phase superpositions, are more sensitive to noise than other configurations. 

We also examined the effect of the training-data distribution by comparing Gaussian and uniform training in the single- and two-component cases and by considering uniform training in the three-component case. 
Overall, the proposed method remains reasonably robust even when the training distribution is less informative than the validation distribution. 
At the same time, the results indicate that the estimation accuracy depends to some extent on the assumed training distribution, and that the degradation becomes more pronounced as the number of superposed components increases. 
This trend suggests that overlap among the component-wise parameter distributions makes the separation and extraction of individual components progressively more difficult. The present encoder--decoder architecture assumes a fixed and known number of components: the latent space contains four slots per component, and its dimension is specified before training. Consequently, the current model cannot infer an unknown or variable number of components, or directly accommodate a signal whose model order differs from that assumed in training. This restriction is appropriate for the present controlled benchmark, but it limits direct application to real data unless independent information supplies the model order. Extending the framework to unknown or variable model order is an important direction for future work.

As an illustrative theoretical reference, we evaluated the local Cram\'{e}r--Rao lower bounds (CRLBs) for the two-component Gaussian-noise settings at the mean parameter values listed in Tables~\ref{tab:case1_parameter-table} and \ref{tab:case2_parameter-table}, where
\begin{equation}
\begin{split}
y_j&=x(t_j;\boldsymbol{\theta})+n_j,
\\
\boldsymbol{\theta}&=(f_0,\phi_0,\tau_0,A_0,f_1,\phi_1,\tau_1,A_1),
\end{split}
\end{equation}
where $n_j$ denotes the additive-noise sample at $t_j$; the noise samples are assumed to be independent and identically distributed (i.i.d.) according to a Gaussian distribution with zero mean and variance $\sigma_{\rm{noise}}^2$.
For $N_t=1000$ samples acquired at $f_s=200$~MHz, the Fisher information matrix is
\begin{equation}
F_{ab}=
\frac{1}{\sigma_{\rm{noise}}^2}
\sum_{j=0}^{N_t-1}
\frac{\partial x(t_j;\boldsymbol{\theta})}{\partial\theta_a}
\frac{\partial x(t_j;\boldsymbol{\theta})}{\partial\theta_b},
\end{equation}
and the local lower bound on the standard deviation of parameter $\theta_a$ is
\begin{equation}
\sigma^{\rm{CRLB}}_{\theta_a}=
\sqrt{\left(F^{-1}\right)_{aa}}.
\end{equation}
For the parameter ordering $(f\,[\mathrm{MHz}],\phi\,[\mathrm{rad}],\tau\,[\mu\mathrm{s}],A)$, the resulting bounds are \\ 
$(0.055,0.479,0.425,0.469)$ and $(0.0062,0.092,0.156,0.409)$ for components 0 and 1 of \textbf{2G-sub}, respectively, 
and $(0.0033,0.035,0.0466,0.124)$ and $(0.0025,0.0308,0.0432,0.116)$ for \textbf{2G-opp}. These values provide local theoretical reference scales for the specified signal model. Because the learned encoder--decoder may be biased and incorporates information from the training distribution, we do not interpret this local CRLB calculation as an efficiency comparison with the autoencoder. A sample-wise CRLB analysis over the full parameter distribution remains beyond the scope of the present work.

Detailed comparisons with other machine-learning-based methods, such as that of Xie's group~\cite{ref:DataDrivenDamped}, are also beyond the scope of the present paper.

More broadly, the present method is promising for short-duration signals in noisy environments. 
In a follow-up study, we further investigate applications of the autoencoder to black hole ringdown gravitational-wave analysis~\cite{ref:Iida2026-3}. 
Ultimately, we aim to apply the proposed approach to black hole ringdown gravitational-wave data and explore its potential for black hole spectroscopy~\cite{Berti:2025hly}.

\section*{Acknowledgements}
This research was supported in part by the Japan Society for the Promotion of Science (JSPS) Grant-in-Aid for Scientific Research [Nos.\ 22K03639 and 26K21813] (H.\ Motohashi) and [Nos.\ 23H01176, 23K25872 and 23K22499] (H.\ Takahashi). This research was supported by the Joint Research Program of the Institute for Cosmic Ray Research, University of Tokyo, and Tokyo City University Prioritized Studies and Research Equipment Program.

\printbibliography

\end{document}